\documentclass[10pt,twocolumn,letterpaper]{article}

\usepackage{iccv}
\usepackage{times}
\usepackage{epsfig}
\usepackage{graphicx}
\usepackage{amsmath}
\usepackage{amssymb}
\usepackage{url}
\usepackage{booktabs}
\usepackage[]{algorithm}
\usepackage[]{algpseudocode}

\makeatletter
\def\BState{\State\hskip-\ALG@thistlm}
\makeatother

\iccvfinalcopy % *** Uncomment this line for the final submission

\def\iccvPaperID{2214} % *** Enter the ICCV Paper ID here

\begin{document}
	% https://www.overleaf.com/5925279985fppbwjrnzksj
	%%%%%%%%% TITLE
	% \title{Are you Certainly Looking at me?- Gradient based Certainty Explanation for Attention}
	% \title{GCA: Gradient-based Certainty Attention Maps for VQA}
	% \title{ Certain Visual Explanation for VQA}
	% \title{Certain Visual Explanation for VQA via Gradient-based Uncertainty Localization}
	% \title{ Certain-CAM: Visual Explanation for VQA via Gradient-based Localization}
	% \title{Certain-CAM: Gradient-based Certain-Class Activation Map for Visual Explanation}
	% \title{Certain-CAM: Gradient-based Certain-Class Activation Map for VQA}
	% \title{Certain-CAM: Gradient based Certain Class Activation Maps}
	\title{U-CAM: Visual Explanation using Uncertainty based Class Activation Maps}
	
	\author{ {Badri N. Patro}  \quad {Mayank Lunayach} \quad {Shivansh Patel}  \quad {Vinay P. Namboodiri} \\
		% \author{ $\textbf{Badri N. Patro}$  \quad $\textbf{Mayank Lunayach}$ \quad $\textbf{Shivansh Patel} $ \quad  $\textbf{Vinay P. Namboodiri}$ \\
		% \author{Badri N. Patro \and Mayank Lunayach \and Shivansh Patel\and Vinay P. Namboodiri\\
		Indian Institute of Technology, Kanpur \\
		{\tt\small \{ badri,mayankl,shivp,vinaypn \}@iitk.ac.in}
		% \and
		% {\tt\small \{ badri,mayankl,shivp,vinaypn \}@iitk.ac.in}
	}
	
	\maketitle
	%\thispagestyle{empty}
	
	%%%%%%%%% ABSTRACT
	\begin{abstract}
		Understanding and explaining deep learning models is an imperative task. Towards this, we propose a method that obtains gradient-based certainty estimates that also provide visual attention maps. Particularly, we solve for visual question answering task. We incorporate modern probabilistic deep learning methods that we further improve by using the gradients for these estimates. These have two-fold benefits: a) improvement in obtaining the certainty estimates that correlate better with misclassified samples and b) improved attention maps that provide state-of-the-art results in terms of correlation with human attention regions. The improved attention maps result in consistent improvement for various methods for visual question answering. Therefore, the proposed technique can be thought of as a recipe for obtaining improved certainty estimates and explanation for deep learning models. We provide detailed empirical analysis for the visual question answering task on all standard benchmarks and comparison with state of the art methods. 
	\end{abstract}
	\vspace{-1.0em}
	%%%%%%%%%%%%%%%%%%%%%%%%%%%%%%%%%%%%%%%%%%%%%%%%%%%%%%%%%%%%%%%%%%%%%%%%%%%%%%%%%%%%%%%%%%%%%%%%%%%%%%%%%%%%%%%%%%%%%%%%%%%%
	\section{Introduction}
	To interpret and explain the deep learning models, many approaches have been proposed. One of the approaches uses probabilistic techniques to obtain uncertainty estimates, \cite{Gal_ICML2016, Gal_NIPS2016}. Other approaches aim at obtaining visual explanations through methods such as Grad-CAM \cite{Das_ICCV2017} or by attending to specific regions using hard/soft attention.
	% Through this work, we aim to obtain both the advantages better than previously proposed such techniques. 
	With the recent probabilistic deep learning techniques by Gal and Ghahramani~\cite{Gal_ICML2016}, it became feasible to obtain uncertainty estimates in a computationally efficient manner. This was further extended to data uncertainty and model uncertainty based estimates~\cite{Kendall_NIPS2017}. Through this work, we focus on using gradients uncertainty losses to improve attention maps while also enhancing the explainability leveraging the Bayesian nature of our approach. The uncertainties that we use are aleatoric and predictive~\cite{Kendall_CVPR2018}.
	\begin{figure}[t]
		\centering
		\includegraphics[width=0.45\textwidth,height=1.0\textheight,keepaspectratio]{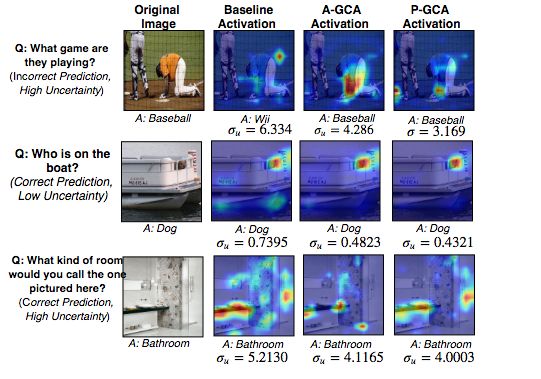}
		\vspace{-0.2cm}
		\caption{The figure shows the activation maps for baseline (MCB~\cite{Fukui_arXiv2016}) and our models (A-GCA and P-GCA). In the first example, the baseline model had predicted the wrong answer and had high uncertainty in prediction. ($\sigma_u$ denotes uncertainty, see Section~\ref{modeling}). Our model gave a correct answer while also minimizing the uncertainty (thus leading to an improved visual explanation). 
			% 		We provide certainty activation map for baseline  model, and our models(A-GCA, P-GCA). In the first raw baseline model predict wrong answer with high uncertainty , due to it confuse about the game.By minimising uncertainty in the image we provide better explanation.
		}
		\vspace{-1.7em}
		\label{fig:intro}
	\end{figure}
	% In this paper, we show that by using gradient with respect to the aleatoric and predictive uncertainty~\cite{Kendall_CVPR2018} losses results in an improvement in terms of obtaining uncertainty estimates that correlate better with misclassification. 
	
	For the estimated uncertainties, we calculate gradients using the approach similar to gradient-based class activation maps~\cite{Das_ICCV2017}. This provides ``certainty maps'' which helps in attending to certain regions of the attention maps. Doing this, we report an improvement in attention maps. This is illustrated in the Figure~\ref{fig:intro}. 
	
	% The proposed method is able to obtain better attention regions while providing Bayesian uncertainty estimates.
	
	Our method combines techniques from both the explanation~\cite{Das_ICCV2017} and uncertainty~\cite{Kendall_NIPS2017} estimation techniques to obtain improved results. We have provided an extensive evaluation. We show that the results obtained for uncertainty estimates show a strong correlation with misclassification, i.e., when the classifier is wrong, the model is usually uncertain. Further, the attention maps provide state of the art correlation with human-annotated attention maps. We also show that on various VQA datasets, our model provides results comparable to the state of the art while significantly improving the performance of baseline methods on which we incorporated our approach. Our method may be seen as a generic way to obtain Bayesian uncertainty estimates, visual explanation, and as a result, improved accuracy for Visual Question Answering (VQA) task.

	Our contributions, therefore, lie in, a) unifying approaches for understanding deep learning methods using uncertainty estimate and explanation b) obtaining visual attention maps that correlate best with human attention regions and c) showing that the improved attention maps result in consistent improvement in results. This is particularly suited for vision and language-based tasks where we are interested in understanding visual grounding, i.e., for instance, if the answer for a question is `dog' (Corresponding question: `Who is on the boat?'), it is important to understand whether the model is certain and whether it is focusing on the correct regions containing a dog. This important requirement is met by the proposed approach.
	
	% For the data uncertainty, we use the combined input (image + question) to capture it. 

\begin{figure}[ht]
		\vspace{-0.2cm}
		\includegraphics[width=0.45\textwidth,height=1.95cm]{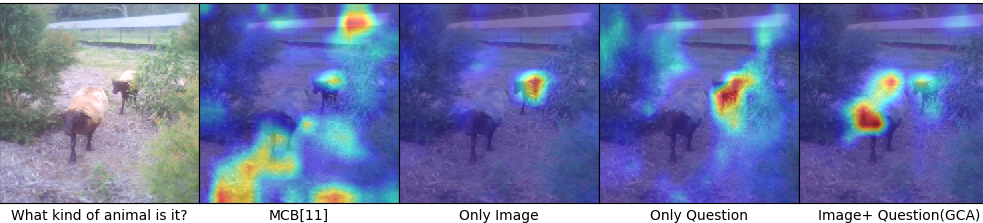}
		% 	\vspace{-0.8cm}
		\caption{The first column is the original image. 2nd, 3rd, 4th, and 5th columns represent the baseline attention, attention when only image uncertainty was minimized, attention when only question uncertainty was minimized, attention when both image and question uncertainties were minimized (proposed model) respectively. }
		\vspace{-0.7cm}
		\label{fig:a1}
	\end{figure}

	% \textbf{Does uncertainty matter in multi-modality?}
	\textbf{Data uncertainty in a multi-modal setting,}
	% We tried with minimizing uncertainties of image and question inputs alone. Minimizing combined image-question embedding uncertainties helps in tapping to more certain regions (and hence better attention maps). 
	Uncertainty in VQA task is two-fold.
	In the example below, Question, "\textit{Which kind of animal is it?}" when asked (irrespective of image), may not be concretely answered. Also, seeing the image alone, in the given setting, the animal (especially the one behind) could easily be mis-classified as a dog or some other animal. These kinds of data uncertainties are tapped \& hence minimized best when we consider uncertainties of the fused input (image+question). In Figure~\ref{fig:a1}, we show the resultant attention maps of baseline (not minimizing uncertainty) \& when we tried to minimize only-image, only-question \& the fused uncertainty respectively.
	%To summarize, through this paper we provide the following contributions :
	%\begin{itemize}
	%    \item We propose a novel gradient-based certainty method for explaining and obtaining better attention maps based on visual explanation in VQA.
	%    \item We show that capturing and minimizing aleatoric and predictive uncertainty using Gaussian cross entropy and variance minimizing loss improves accuracy. 
	%  \item We provide various comparisons and results to show that we obtain better attention maps that correlate well with human attention maps and outperform other techniques for VQA. Further, we show that obtaining better attention maps also aids in obtaining better accuracies while solving for VQA.  A detailed empirical analysis for the same is provided.
	%  \item Improving overall accuracy by 1.1\% by minimizing uncertainty for the classification task. Estimation \& visualization of the aleatoric  \& predictive uncertainty for the classification task.
	%\end{itemize}
	%  \begin{figure}
	% 	\centering
	% 	\includegraphics[width=0.45\textwidth]{fig/ICCV_motivation.png}
	% 	\caption{Illustration of main figure.}
	% 	\label{fig:intro1}
	% \end{figure}
	%%%%%%%%%%%%%%%%%%%%%%%%%%%%%%%%%%%%%%%%%%%%%%%%%%%%%%%%%%%%%%%%%%%%%%%%%%%%%%%%%%%%%%%
	\section{Related work}
	% \label{sec:lit_surv}
	%%%%%%%%%%%%%%%%%%%%%%%%%%%%%%%%%%%%%%%%%%%%%%%%%%%%%%%%%%%%%%%%%%%%%%%%%%%%%%%%%%%%%%%
	The task of Visual question answering~\cite{Malinowski_NIPS2014, VQA, Ren_NIPS2015, Goyal_CVPR2017, Noh_CVPR2016} is well studied in the vision and language community, but it has been relatively less explored for providing explanation\cite{selvaraju2017grad} for answer prediction. Recently, lot of works that focus on explanation models, one of that  is image captioning for provide basic explanation for image ~\cite{Barnard_JMLR2003,Farhadi_ECCV2010,Kulkarni_CVPR2011,Socher_TACL2014,Vinyals_CVPR2015,Karpathy_CVPR2015,Xu_ICML2015,Fang_CVPR2015,Chen_CVPR2015,Johnson_CVPR2016,Yan_ECCV2016}.  ~\cite{Patro_EMNLP2018MDN}  has proposed an exemplar-based explanation method for generating question based on the image. Similarly, ~\cite{Patro_COLING2018learning} has suggested a discriminator based method to obtain an explanation for paraphrase generation in text.  In VQA, \cite{Zhu_CVPR2016}\cite{Xu_ECCV2016} have proposed interesting methods for improving attention in the question. %An interesting work obtains a varied set of modules for answering questions of different types \cite{andreas16naacl}.
	Work that explores image and question jointly and is based on hierarchical co-attention is ~\cite{Lu_NIPS2016}. ~\cite{Shih_CVPR2016, Yang_CVPR2016, Li_NIPS2016, Patro_CVPR2018dvqa} have proposed attention-based methods for the explanation in VQA, which use question to attend over specific regions in an image. ~\cite{Fukui_arXiv2016, Kim_ICLR2017,kim_NIPS2018bilinear} have suggested exciting works that advocate multimodal pooling and obtain close to state of the art in VQA. ~\cite{Patro_CVPR2018dvqa} has proposed an exemplar-based explanation method to improve attention in VQA. We can do systematic comparison of image-based attention while correlating with human attention maps as shown by \cite{Das_EMNLP2016}. Other methods~\cite{kurmi2019curriculum,kurmi2019looking} explore the correlation between the distributions of the dataset. The computational efficiency of these deep learning models can be improved by~\cite{Singh_2019_CVPR}.
	% Also, we need to ensure that our approach works in other dataset  distributions~\cite{kurmi2019curriculum,kurmi2019looking} of images while taking less time ~\cite{Singh_2019_CVPR} to train our explanation model.
	
	%In our approach, We proposed a gradient-based certainty explanation mask which minimises uncertainty in various attention regions to improve answer prediction probability in VQA. Interestingly, we show that by combining this with the proposed method it further improves our results.
	
	Recently a lot of researchers have focused on estimating uncertainty in the deep models. \cite{blundell_ICML2015weight} has first proposed a method to learn uncertainty in the weights of the neural network. Kendall \textit{et.al.} ~\cite{kendall2015bayesian} has proposed method to measure model uncertainty for image segmentation task. They observed that softmax probability function approximates relative probability between the class labels, but does not provide information about the model's uncertainty. The work by ~\cite{Gal_ICML2016, Fortunato_Arxiv2017} estimates model uncertainty of the deep network (CNN, RNN) with the help of dropout~\cite{srivastava2014dropout}. \cite{teye2018bayesian} has estimated uncertainty for batch normalized deep networks. \cite{Kendall_NIPS2017, Kendall_CVPR2018, smith2018understanding} have mainly decomposed predictive uncertainty into two major types, namely aleatoric and epistemic uncertainty, which capture uncertainty about the predicted model and uncertainty present in the data itself. \cite{malinin2018predictive} suggested a method to measure predictive uncertainty with the help of model and data uncertainty. Recently, ~\cite{kurmi_cvpr2019attending} proposed a certainty method to bring two data distributions close for the domain adaption task. Here, our objective is to analyze and minimize the uncertainty in attention mask to predict answer in VQA. In our approach, We are proposing a gradient-based certainty explanation mask which minimizes uncertainty in attention regions to improve the correct answer's predicted probability in VQA. %Interestingly, we show that by combining this with the proposed method it further improves our results.%We proposed a gradient-based certainty explanation mask which minimizes uncertainty and improves attention.
	Our method also provides visual explanation based on uncertainty class activation maps, capturing and visualizing the uncertainties present in the attention maps in VQA.
	%%%%%%%%%%%%%%%%%%%%%%%%%%%%%%%%%%%%%%%%%%%%%%%%%%%%%%%%%%%%%%%%%%%%%%%%%%%%%%%%%%%%%%%%%%
	\section{Modeling Uncertainty}\label{modeling}
	%%%%%%%%%%%%%%%%%%%%%%%%%%%%%%%%%%%%%%%%%%%%%%%%%%%%%%%%%%%%%%%%%%%%%%%%%%%%%%%%%%%%%%%
	% We consider two types of uncertainty explaination to model  uncertainty present in attention network, one due to uncertainty present in the data (Aleatoric), and the other due to model (Epistemic uncertainty).
	We consider two type of uncertainties present in the deep network, one due to uncertainty present in the data (Aleatoric), and the other due to model (Epistemic uncertainty).
	\subsection{Modeling Aleatoric Uncertainty}
\begin{figure}
		\centering
		\includegraphics[width=0.45\textwidth]{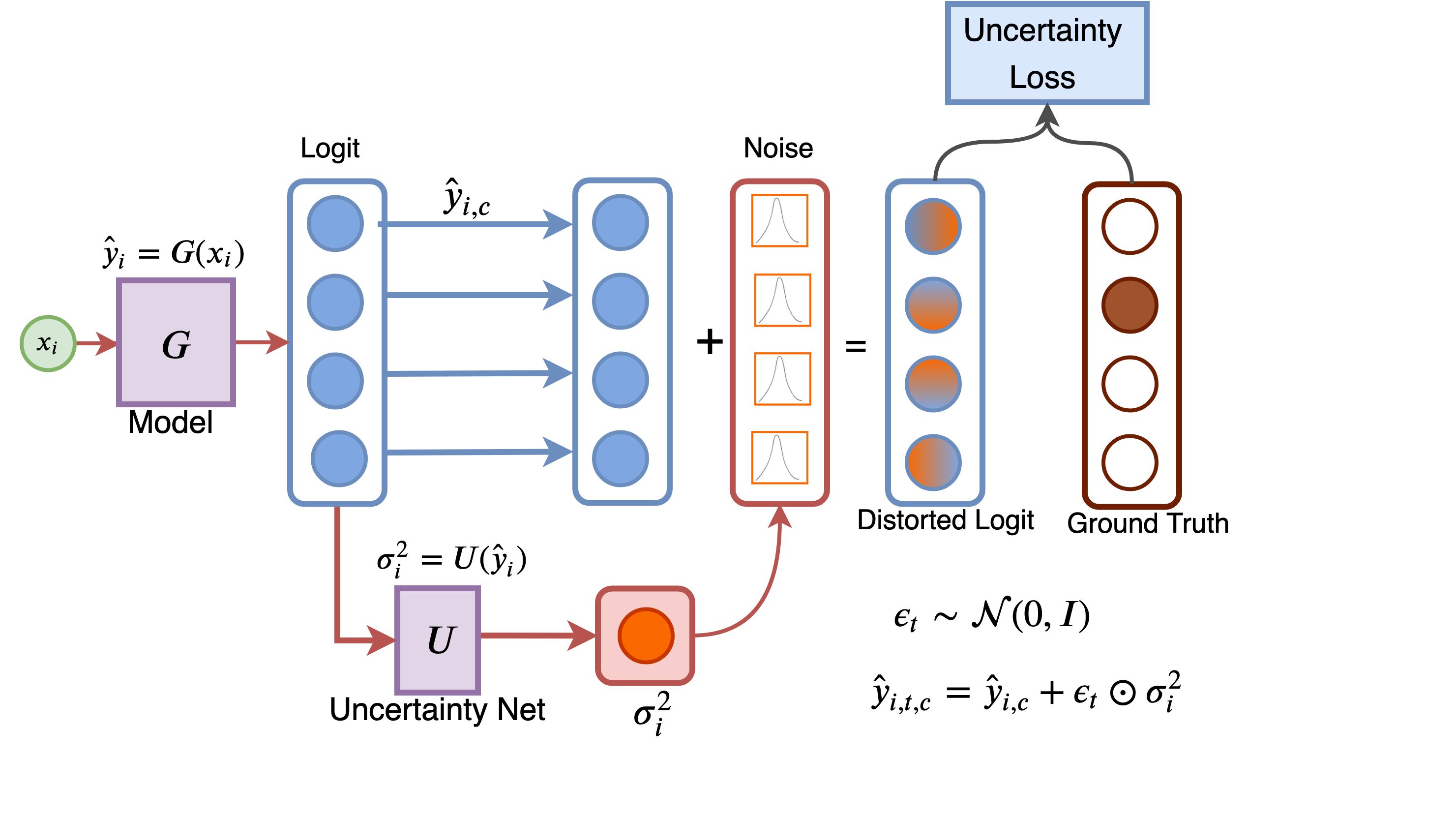}
		\vspace{-0.4cm}
		\caption{Illustration of Uncertainty Loss}
		\label{fig:basic}
		\vspace{-1em}
	\end{figure}

	Given an input $x_i$ the model ($G$) predicts the logit output $\hat{y}_i$ which is then an input to uncertainty network ($U$) for obtaining the variance $\sigma_i^2$ as shown in Figure-\ref{fig:basic}. To capture Aleatoric uncertainty~\cite{Kendall_NIPS2017}, we learn the observational noise parameter $\sigma_{i}$ for each input point $x_i$. Then, Aleatoric uncertainty, $(\sigma^2_{a})_{i}$ is estimated by applying softplus function on the output logit variance. This is given by,
	\begin{equation}
	\label{aleatoric_uncertianty}
	(\sigma^2_{a})_{i}=Softplus(\sigma^2_i)= log(1+exp(\sigma^2_i))
	\end{equation}
	For calculating the aleotoric uncertainty loss, we perturb the logit value ($y_{i}$) with Gaussian noise of variance $(\sigma^2_{a})_{i}$ (diagonal matrix with one element corrosponding to each logits value) before the softmax layer. The logits reparameterization trick~\cite{Kendall_CVPR2018} and ~\cite{Gal2016Uncertainty} combines $\hat{y}_{i,c}$ and $\sigma_i$  to give $\mathcal{N}(\hat{y}_{i,c},\sigma_i^2)$. We then obtain a loss with respect to ground truth. It is expressed as:
	\begin{equation}
	\label{aleatoric_variance}
	\hat{y}_{i,c,t}=y_{i,c}+\epsilon_{t}*\sigma^2_{i}, where \quad \epsilon_{t} \sim \mathcal{N}(0,I)
	\end{equation}
	\begin{equation}
	\label{l6}
	\mathcal{L}_a= \sum_{i}\log \frac{1}{T}\sum_{t}{\exp{(\hat{y}_{i,c,t} - \log \sum_{c^`}\exp{\hat{y}_{i,c^{`},t}})}}
	\end{equation}
	where $\mathcal{L}_a$ is the aleatoric uncertainty loss (AUL), T is the number of Monte Carlo  simulations. $c^{'}$ is a the class index of the logit vector $y_{i,t}$ which is defined for all the classes.
	%$\sigma_{i,c}$ is the standard deviation, ( $\sigma_{i,c}=\sqrt{v_{i,c}}$). 
	% The classifier is trained by jointly minimizing both the classification loss, $\mathcal{L}_y$ and the aleatoric loss, $\mathcal{L}_a$. 
	
	\subsection{Modeling Predictive Uncertainty}
	To obtain the model uncertainty, we measure epistemic uncertainty. However, estimating epistemic uncertainty~\cite{malinin2018predictive} is computationally expensive, and thus we measure the predictive uncertainty, having both aleatoric and epistemic uncertainties present in it. To estimate it, we sample weights in the Bayesian networks $G$ and then perform Monte Carlo simulations over the model to obtain the predicted class probabilities $p(y_{i,t})$. That is, 
	\begin{equation*}
	\label{eq4}
	\begin{split}
	&O(\hat{y}_{i,t})= G^t(x_{i}) \quad v^{a}_{i,t}= Softplus(U^t(\hat{y}_{i,t}))\\
	&p(\hat{y}_{i,c}|x_i,X_I)=(\frac{1}{T} \sum_{t=1}^{T}{Softmax{ O(\hat{y}_{i,t})}})_c
	\end{split}
	\end{equation*}
	where c is the answer class, $G^t \sim G,  U^t \sim U$  and $ v^{a}_{i,t}$ is the aleatoric variance of each logit in the $t^{th}$ MC Simulation. The entropy of the sampled logit's probabilities can be calculated as:
	\begin{equation}
	\label{eq5}
	H(\hat{y}_{i})= -\sum_{c=1}^{C}{p(\hat{y}_{i,c})*\log{p(\hat{y}_{i,c}})}
	\end{equation}
	The predictive uncertainty contains entropy and aleatoric variance when it's expectation is taken across $T$ number of Monte Carlo simulations:
	\begin{equation}
	\label{eq6}
	\sigma^2_p =H(\hat{y}_{i}) + \frac{1}{T}\sum_{t=1}^{T} {v^{a}_{i,t}}
	\end{equation}
	where $H(\hat{y}_{i})$ is the entropy of the probability $p(\hat{y}_{i})$, which depends on the spread of class probabilities while the variance (second term in the above equation) captures both the spread and the magnitude of logit outputs, $\hat{y}_{i,t}$.  In Equation~\ref{aleatoric_variance}, we can replace $\sigma^2_a$ with predictive uncertainty  $\sigma^2_p$ (mentioned above in Equation~\ref{eq6}) to get the predictive uncertainty loss (PUL). 
	
	\begin{figure*}
		\centering
		\includegraphics[width=0.9\textwidth]{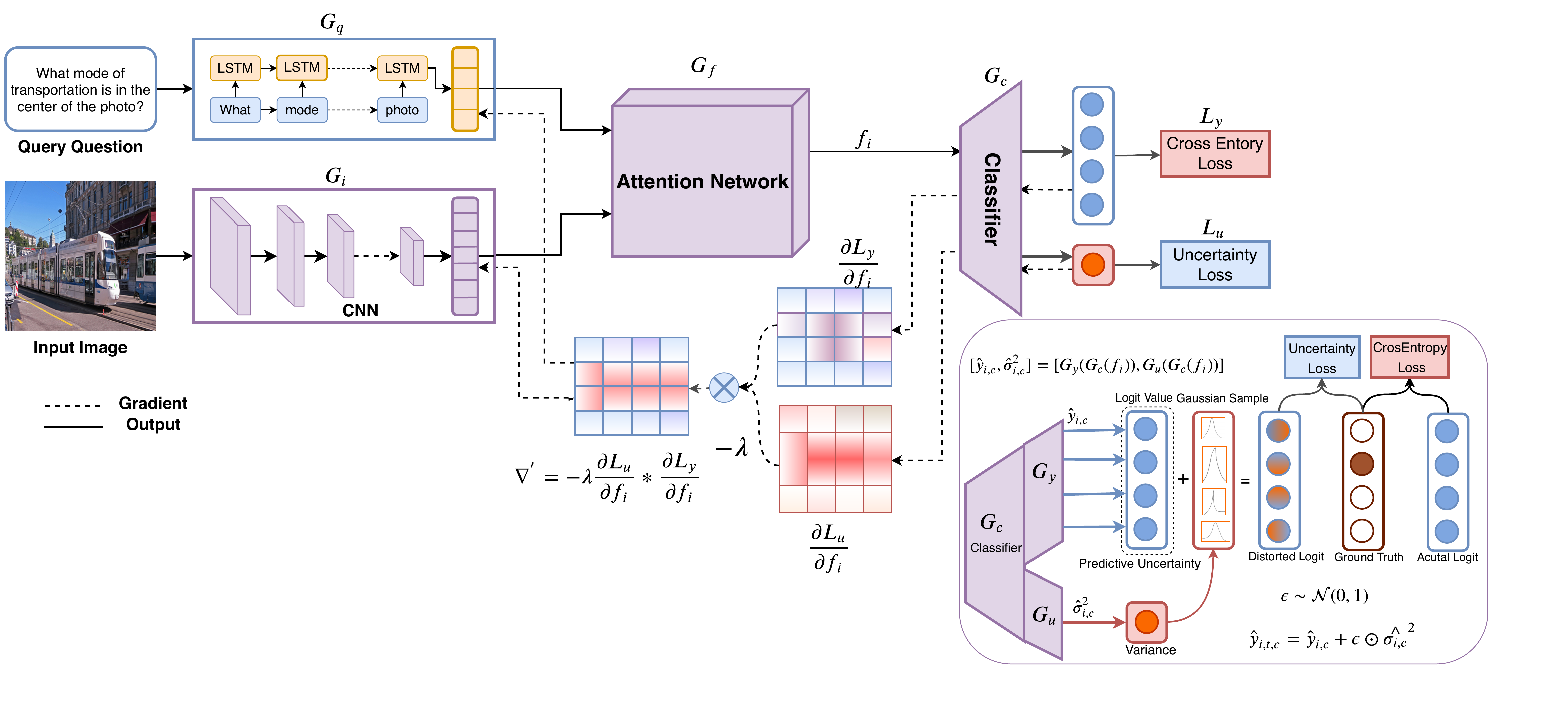}
		\vspace{-0.4cm}
		\caption{Illustration of model Gradient-based Certainty Attention Mask (GCA) and its certainty mask. We obtain image feature and question feature using CNN and LSTM, respectively. We then obtain attention mask using these features, and classification of the answer is done based on the attended feature. %We  have  improved  the  attention  mask  with  the certainty explanation approach.
		}
		\label{fig:main}
		\vspace{-1em}
	\end{figure*}
	%%%%%%%%%%%%%%%%%%%%%%%%%%%%%%%%%%%%%%%%%%%%%%%%%%%%%%%%%%%%%%%%%%%%
	\vspace{-0.51em}
	%%%%%%%%%%%%%%%%%%%%%%%%%%%%%%%%%%%%%%%%%%%%%%%%%%%%%%%%%%%%%%%%%%%%%%%%%%%%%%%%%
	\section{ Method}
	%%%%%%%%%%%%%%%%%%%%%%%%%%%%%%%%%%%%%%%%%%%%%%%%%%%%%%%%%%%%%%%%%%%%%%%%%%%%%%%%%
	\vspace{-0.51em}
	% The main focus in our approach for solving visual question answering (VQA) is to use supervision obtained from visual explanation methods such as Grad-CAM to improve attention. As mentioned earlier, using Grad-CAM as attention shows improved performance in comparison to just using attention alone. Therefore, we believe that Grad-CAM or any other visual explanation method can be used in this setting. Further, by learning both visual explanation and attention jointly in an adversarial setting we observe improvements in both as shown empirically. 
	\textbf{Task}: We solve for VQA~\cite{VQA} task. The key difference in our architecture as compared to the existing VQA models is the introduction of gradient-based certainty maps. A detailed figure of the model is given in the Figure-~\ref{fig:main}. We keep other aspects of the VQA model unchanged. In a typical open-ended VQA task, we have a multi-class classification task. A combined (image and question) input embedding is fed to the model. Then, the output logits are fed to a softmax function, giving probabilities of the predictions in the multiple-choice answer space. That is,
	$\hat{A}=\underset{A \in \Omega}{argmax}P(A|I, Q, \theta)$,
	where $\Omega$ is a set of all possible answers, I is the image, Q is the corresponding question, and $\theta$ is representing the parameters of the network. 
	%This module focuses on minimizing uncertainty in the answer classification module. We minimize uncertainty in two ways for visual question answer task. First is uncertainty present in the data that is aleatoric uncertainty and the second one is predictive uncertainty that captures uncertainty in the model and data as well.
	
	%In VQA task is defined as given image and natural question based on the image the model construct a joint feature vector for predict answer probability. In  this  paper  we  adopt  a  classification  framework  that uses the image embedding combined with the question embedding to solve for the answer using a softmax function in a multiple choice setting as mentioned like:
	%$\hat{A}=\underset{A \in \Omega}{argmax}P(A|I,Q : \theta)$,
	%where $\Omega$ is a set of all possible answers and $\theta$ represents the parameters in the network.
	% \vspace{-1em}
	
	%%%%%%%% 
	%  \begin{figure}
	% 	\centering
	% 	\includegraphics[width=0.45\textwidth]{latex/fig/iccv_complete.jpg}
	% 	\caption{Illustration of Aleatoric loss.}
	% 	\label{fig:ale_loss}
	% \end{figure}
	% \subsection{U-CAM Approach}
	\subsection{U-CAM Approach}
	The three main parts of our method are Attention Representation, Uncertainties Estimation, and computing gradients of uncertainty losses. In the following sections, we explain them in detail.
	\vspace{-1em}
	\subsubsection{Attention Representation}
	\vspace{-0.7em}
	We obtain an embedding, $g_i \in \mathcal{R}^{u \times v \times C}$ where u is width, v is height of the image and C represents the number of applied filters on the image $X_i$ in the convolution neural network (CNN). The CNN is parameterized by a function $G_i(X_i,\theta_i)$, where $\theta_i$ represents the weights. Similarly, for the query question $X_Q$, we obtain a question feature embedding $g_q$ using a LSTM network. This network is parameterized by a function $G_q(X_q,\theta_q)$, where $\theta_q$ represents the weights. Both $g_i$ and $g_q$  are fed to an attention network that combines the image and question embeddings using a weighted softmax function and produces a weighted output attention vector, $g_f$ as illustrated in Figure~\ref{fig:main}. Various kinds of attention networks have been proposed in the literature.  In this paper, we tried with SAN~\cite{Yang_CVPR2016} and MCB~\cite{Fukui_arXiv2016}. %The attention mechanism works as follows:
	%\begin{equation}
	%    \begin{split}
	%        & g_{a}=W_{a}\tanh(W_{i}g_i+ W_{q}g_q +b_{c})\\
	%        & \alpha = \text{Softmax}(g_{a}) \\
	%        & g_{f} = \sum_{}^{}\sum_{}^{}{\alpha*g_i} \\
	%        % & g_{out} = g_{att} + g_q
	%    \end{split}
	% \vspace{-2em}
	%\end{equation}
	%Where, $W_{(.)},b_{(.)}$ are the learn-able weights and bias in the attention network.
	% \vspace{-1.3em}
	% \textbf{Bayesian Classifier}
	Finally, we obtain attention feature $f_i$ using attention extractor network $G_f:f_i=G_f (g_i,g_q)$.
	%  , where $g_i$ and $g_q$ are input image and question embeddings respectively.
	The attended feature $f_i$ is passed through a classifier and the model is trained using the cross-entropy loss.
	Many a times, model is not certain about the answer class to which the input belongs, which sometimes leads to decrease in accuracy. To tackle this, we have proposed a technique to reduce the class uncertainty by increasing the certainty of the attention mask. Additionally, we also incorporate a loss based on the uncertainty which is described next. %Many a times, model is not certain about the answer class to which the input belongs, which sometimes leads to decrease in accuracy. To tackle this, we have proposed a technique to reduce the class uncertainty by increasing the certainty of the attention mask. We have adopted a Bayesian framework~\cite{Kendall_NIPS2017} to estimate the uncertainty of the answer efficiently. We make our answer classifier Bayesian and perform probabilistic inference for obtaining the final answer probability. To achieve this, we apply dropout after every fully connected (FC) layer in the classifier thus  making a Bayesian model \cite{Gal_NIPS2016}, and the whole network is trained with the help of both the uncertainty loss (More details are present in the Section \ref{costfun_ale}) \& the answer classification (cross-entropy) loss. The uncertainty loss shall help in making the attention map more robust \& certain, which ultimately leads to an improvement in the answer prediction. 
	% To verify, we measure the predictive uncertainty for our final prediction and found it to be lower. 
	\vspace{-1em}
	\subsubsection{Estimating Uncertainties: Aleatoric \& Predictive}
	\vspace{-0.7em}
	
	% We consider two types of uncertainties to explain our attention network, one due to uncertainty present in the data (Aleatoric), and the other due to model (Epistemic uncertainty).
	
	The attention feature, $f_i$ obtained from the previous step is fed to the classifier $G_c$. The output of the classifier is fed to $G_y$, which produces class probabilities, $y_i$. $G_c$'s output is also fed to a variance predictor network, $G_v$, which outputs the logits' variance, $\sigma_{i}$ as mentioned in the Equation~\ref{aleatoric_uncertianty}. For calculating the aleatoric uncertainty loss, we perturb the logit value ($y_{i}$) with Gaussian noise of variance $(\sigma^2_{a})_{i}$
	%(diagonal matrix with one element corresponding to each logits value) 
	before the softmax layer. The Gaussian likelihood for classification is given by $p(y_i|f_i,w)=\mathcal{N}(y_i;G_y(G_c(f_i;w)),{\tau}^{-1}(f_i;w))$, where   $w$ represents model's parameters, $\tau$ is the precision, $f_i$ is the attended fused input, and $G_y(G_c(.))$ is the output logit producing network as shown in the Figure-~\ref{fig:main} . The above setting represents the perturbation of model output with the variance of the observed noise, $\tau^{-1}$. We make sure that ${\tau}(.)$  is a positive or positive definite matrix (in case of Multivariate) by using the logit reparameterization trick ~\cite{Kendall_CVPR2018,Gal2016Uncertainty}. Finally, we then obtain an aleatoric loss, $\mathcal{L}_a$ with respect to ground truth as mentioned in the Equation-~\ref{l6}.   
	Our proposed model, which uses this loss as one of the components of its uncertainty loss, is called Aleatoric-GCA (A-GCA). Along with aleatoric loss $\mathcal{L}_a$, we combine $\mathcal{L}_{VE}$ and $\mathcal{L}_{UDL}$ as mentioned in the Equation~\ref{ve} and \ref{udl} respectively to get total uncertainty loss $\mathcal{L}_u$. The classifier is trained by jointly minimizing both the classification loss, $\mathcal{L}_y$ and the uncertainty loss, $\mathcal{L}_u$. In Equation~\ref{aleatoric_variance}, we can replace $\sigma^2_a$ with predictive uncertainty  $\sigma^2_p$ (mentioned above in Equation-~\ref{eq6}) to get the predictive uncertainty loss (PUL). Accordingly, the model which uses this loss as one of the constituents of its uncertainty loss is called Predictive-GCA (P-GCA).  
	% https://www.quora.com/Which-is-the-best-measure-of-uncertainty-variance-or-entropy-or-are-they-both-equivalent 
	Next, we compute the gradients of standard classification loss and uncertainty loss with respect to attended image feature, $f_i$.  Besides training, we also use these gradients to obtain visualizations describing important regions responsible for answer prediction, as mentioned in the qualitative analysis section (Section~\ref{ResultQual}).
	% \begin{figure}
	% 	\centering
	% 	\includegraphics[width=0.45\textwidth]{fig/uncertainty_explanation.png}
	% 	  \vspace{-1.52em}
	% 	\caption{A closer look at the training procedure.}
	% 	\label{fig:detail}
	% 	  \vspace{-1.52em}
	% \end{figure}
	\vspace{-1.2em}
	\subsubsection{Gradient Certainty Explanation for Attention}\label{certainit_variants}
	Uncertainty present in the attention maps often leads to uncertainty in the predictions and can be attributed to the noise in data and the uncertainty present in the model itself. We improve the certainty in these cases by adding the certainty gradients to the existing  Standard Cross-Entropy (SCE) loss gradients for training the model during backpropagation.
	
	% enhancing the gradient of the attention regions based on the gradient of the certainty regions. 
	% The uncertainty estimation of the attention network helps in deciding which regions in the input are certain.
	Our objective is to improve the model's attention in the regions where the classifier is more certain. 
	% by promoting positive transfer. 
	The classifier will perform better by focusing more on certain attention regions, as those regions are more suited for the classification task.
	% \textbf{Aleatoric Gradient Certainty for Attention:(A-GCA)}
	We can get an explanation for the classifier output as done in the existing Grad-CAM approach ($\frac{\partial{\mathcal{L}_y}}{\partial{f_i}}$). But that explanation does not take the model and data uncertainties into the account. We improve this explanation using the certainty gradients ($-\frac{\partial{\mathcal{L}_u}}{\partial{f_i}}$).
	%  The uncertainty in  VQA basically occurs due to the confusion present in the image or question feature representation (aleatoric) or noise in the model parameters (epistemic) as mentioned in Figure-~\ref{fig:a1}.
	If we can minimize uncertainty in the VQA explanation, then uncertainties in the image and question features, and thus uncertainties in the attention regions would be subsequently reduced. It is the uncertain regions which are a primary source for errors in the prediction, as shown in Figure~\ref{fig:intro}. 
	
	In our proposed method, 
	% $\alpha_{gca} \in  \mathcal{R}^{u \times v}$ of width u and height v,
	we compute the gradient of the Standard Classification (cross entropy) loss $L_{y}$ with respect to attention feature i.e. $\frac{\partial{L_y}}{\partial{f_i}}$ and also the gradient of the uncertainty loss $\mathcal{L}_u$ i.e. $\frac{\partial{\mathcal{L}_u}}{\partial{f_i}}$. The obtained uncertainty gradients are passed through a gradient reversal layer, giving us the certainty gradients, i.e., $-\frac{\partial{\mathcal{L}_u}}{\partial{f_i}}$.
	\begin{equation}
	\nabla_{y}^{'} = -\lambda \frac{\partial{\mathcal{L}_u}}{\partial{f_i}} * \frac{\partial{\mathcal{L}_y}}{\partial{f_i}}\end{equation}
	The positive sign of gradient $\nabla_{y}^{'}$ indicates that the attention certainty is activated on these regions and vice-versa. It can be expressed as:
	\begin{equation}\nabla_{y}^{''} =ReLU(\nabla_{y}^{'})+\gamma ReLU(-\nabla_{y}^{'})\end{equation}
	We apply a ReLU activation function on the product of gradients of the attention map and the gradients of certainty as we are only interested in attention regions that have a positive influence on interested answer class, i.e. attention regions whose intensity should be increased in order to increase answer class probability $y_c$, whereas negative values are multiplied by $\gamma$ (large negative number) as the negative attention regions are likely to belong to other categories in image.  As expected, without this ReLU, localization maps sometimes highlights more than just the desired class and achieves lower localization performance. Then we normalize $\nabla_{y}^{''}$ to get attention regions which are highly activated and giving more weight to certain regions and is expressed as:
	\begin{equation}\nabla_{y}^{'''} =\frac{(\nabla_{y}^{''})_{u,v}}{\sum_{u}{}\sum_{v}{}{(\nabla_{y}^{''})_{uv}}}\end{equation}
	Images with higher uncertainty are equivalent to having lower certainty, so the certain regions of these images should have lower attention values. We use residual gradient connection to obtain the final gradient, which is the sum of gradient mask of $\mathcal{L}_y$ (with respect to attention feature) and the gradient certainty mask $\nabla_{y}^{'''}$ and is given by:
	\begin{equation}\frac{\partial{L_y}}{\partial{f_i}}=\frac{\partial{L_y}}{\partial{f_i}} +\nabla_{y}^{'''} \end{equation}
	% \[\nabla_{y}=\nabla_{y} +\nabla_{y}^{'''}} \]
	Where $\frac{\partial{L_y}}{\partial{f_i}} $ is the gradient mask of $\mathcal{L}_y$ when gradients are taken with respect to attention feature. More details are given in the Algorithm ~\ref{alg:GCA}.
	
	%  We compute gradients of the predictive uncertainty $\sigma_p^2$ of our classifier with respect to the features $f_i$, i.e. $\frac{\partial {\sigma^2} }{\partial {f_i}} $ for the attention regions. 
	
	%  If $\sigma^2$ is $\sigma_p^2$ which denotes Predictive-GCA  (P-GCA) and If $\sigma^2$ is $\sigma_a^2$ which denotes Aleatoric-GCA  (A-GCA) method.
	% Considering both data and model certainty, the Gradient Certainty for  Attention (P-GCA) attention network results in an improved attention network and the attention confidence also increases. 

	%%%%%%%%%%%%%%%%%%%%%%%%%%%%%%%%%%%%%%%%%%%%%%%%%%%%%%%%%%%%%%%%%%%%%%
	%  ref: https://arxiv.org/pdf/1701.05369.pdf
	%  http://www.gatsby.ucl.ac.uk/~ucgtcbl/papers/ForBluVin2017a.pdf
	%  http://galton.uchicago.edu/~eichler/stat22000/Handouts/l12.pdf
	
	% % %%%%%%%%%%%%%%%%%%%%%%%%%%%%%%%%%%%%%%%%%%%%%%%%%%%%%%%%%%%%%%%%%%%%%%%
	% % %  ref: https://arxiv.org/pdf/1701.05369.pdf
	% % %  http://www.gatsby.ucl.ac.uk/~ucgtcbl/papers/ForBluVin2017a.pdf
	% %  http://galton.uchicago.edu/~eichler/stat22000/Handouts/l12.pdf
	\begin{algorithm}
		\caption{Gradient Certainty base Attention (GCA)}\label{alg:GCA}
		\begin{algorithmic}[1]
			\scriptsize
			\Procedure{GCA}{$I,Q$}
			% \BState\emph{ Input,Output}:
			\State {\bfseries Input:}  Image $X_I$, Question $X_Q$
			\State {\bfseries Output:}  Answer $y_c$      
			% \BState\emph{ Aleatoric Loss}:
			\While{loop}
			\State  \text{Attention features  $ G_f(G_i(X_I),G_q(X_Q))\gets f_i$}
			\State  \text{Answer Logit $ G_y(G_c(f_i))\gets \hat{y}$}
			\State  \text{Data Uncertainty $ G_v(G_c(f_i))\gets \sigma^2_A$}
			\If {A-GCA}:
			\State  $\sigma^2_W=\sigma^2_A$
			\ElsIf {P-GCA}:
			\State  $\sigma^2_W=\sigma^2_A + H(\hat{y}_{i,t}) , \text{(Ref: eq-~\ref{eq6})}$
			\EndIf
			\State  \textit{Ans cross entropy $\mathcal{L}_y \gets$ loss$(\hat{y},y)$}
			\State Variance Equalizer  $\mathcal{L}_{VE}: =\sum{ReLU(\exp^{\sigma^2_{w}} - \exp^{I}})\cite{Dorman_github2018} $,
			%  \State Cross Entropy  $\it{L_{CE}}= -{\sum_{}^{} y \log \texttt{p}(\hat{y}|F(.))}$
			\While{$t=1: \# MC-Samples$}
			\State \textit{Sample ${\epsilon}_t^{w} \sim \mathcal{N}(0,\sigma^2_W)$ }
			\State \textit{Distorted Logits:$ \hat{y}_{i,t} ={\epsilon}^{w}_t + \hat{y}_i $}
			\State \textit{Gaussian Cross Entropy $\it{L_{p}}= -{\sum_{}^{} y \log \texttt{p}(\hat{y_d}|F(.))}$}
			\State \textit{Distorted Loss :$\it{\mathcal{L}_{UDL}}= \exp(\it{\mathcal{L}_{y}}-\it{\mathcal{L}_{p}})^2$}\cite{Dorman_github2018}
			\State \textit{Aleatoric uncertainty loss $\it{\mathcal{L}_{u}}= \it{\mathcal{L}_{p}}+\it{\mathcal{L}_{VE}+\mathcal{L}_{UDL}}$}
			\EndWhile
			\State \textit{Compute Gradients w.r.t $f_i$,$\nabla_y= \frac{\partial \mathcal{L}_y}{ \partial  f_i}$, $\nabla_u= \frac{\partial \mathcal{L}_{u}}{ \partial  f_i}$}
			\State \textit{Certainty Gradients $\nabla_u^{'}= -\lambda\nabla_u *\nabla_y$}
			\State \textit{Certainty Activation $\nabla_u^{''}= ReLU(\nabla_u^{'})+ \gamma ReLU(-\nabla_u^{'})$}
			\State \textit{Final Certainty Gradients $\nabla_u^{'''}=softmax(\nabla_u^{''})$}
			\State \textit{Final Attention Gradient $\nabla_y^{}=\nabla_y^{}+\nabla_u^{'''} $}
			\State  \textit{update $\theta_f \gets \theta_f - \eta \nabla_y^{}$} 
			\EndWhile
			\EndProcedure
		\end{algorithmic}
	\end{algorithm}
	% \vspace{-0.5em}
	%%%%%%%%%%%%%%%%%%%%%%%%%%%%%%%%%%%%%%%%%%%%%%%%%%%%%%%%%%%%%%%%%%%%%%%%%%%%%%%%%%%%
	\subsection{Cost Function} \label{costfun_ale}
	% \vspace{-0.5em}
	We estimate aleatoric uncertainty in logits space by perturbing each logit using the variance obtained from data. The uncertainty present in the logits value can be minimized using cross-entropy loss on Gaussian distorted logits, as shown in the Equation~\ref{l6}. The distorted logit is obtained using a Gaussian multivariate function, having positive diagonal variance. To stabilize the training process ~\cite{Gal2016Uncertainty}, we add an additional term to the uncertainty loss, calling it Variance Equalizer(VE) loss, $\mathcal{L}_{VE}$.
	% The loss function in the equation-~\ref{ve} ensures that the model precision is always positive.
	\begin{equation}\label{ve}\mathcal{L}_{VE}=\exp{(\sigma_i^2)}-\exp{({\sigma_0}^2)}\end{equation}
	where $\sigma_0$ is a constant.
	The uncertainty distorted loss (UDL) is the difference between the typical cross-entropy loss and the aleatoric/predictive loss estimated in the Equation~\ref{l6}. The scalar difference is passed  to an activation function to enhance the difference in either direction and is given by :
	%  \[L_{UDL}= -\it{ELU}\{ \mathcal{L}_{p}-\mathcal{L}_{y}\}\]
	\begin{equation}
	\label{udl}
	\mathcal{L}_{UDL}=\begin{cases}
	\alpha (\exp^{[\mathcal{L}_{p}-\mathcal{L}_{y}]}-1), & \text{if $[\mathcal{L}_{p}-\mathcal{L}_{y}]<0$}.\\
	[\mathcal{L}_{p}-\mathcal{L}_{y}], & \text{otherwise}.
	\end{cases}
	\end{equation}
	By putting this constraint, we ensure that the predictive uncertainty loss does not deviate much from the actual cross-entropy loss. The total uncertainty loss is the combination of Aleatoric (or prediction uncertainty loss), Uncertainty Distorted Loss, and Variance equalizer loss.  
	\begin{equation}\mathcal{L}_{u}=\mathcal{L}_{p} + \mathcal{L}_{VE}+ \mathcal{L}_{UDL}\end{equation}
	The final cost function for the network combines the loss obtained through uncertainty (aleatoric or predictive) loss $\mathcal{L}_u$ for the attention network with the cross-entropy.
	
	The cost function used for obtaining the parameters $\theta_f$ of the attention network, $\theta_c$ of the classification network, $\theta_y$ of the prediction network and $\theta_u$ for uncertainty network is as follows:
	\[C(\theta_f,\theta_c,\theta_y,\theta_u)=\frac{1}{n}\sum_{j=1}^{n}{ L^j_y(\theta_{f},\theta_c,\theta_y)} + \eta L^j_u(\theta_{f},\theta_c,\theta_u)\]
	where n is the number of examples, and $\eta$ is the hyper-parameter which is fine-tuned using validation set, $L_y$ is standard cross-entropy loss and $L_u$ is the uncertainty loss.
	We train the model with this cost function until it converges so that the parameters. $(\hat{\theta}_f,\hat{\theta}_c,\hat{\theta}_y,\hat{\theta}_u)$ deliver a saddle point function
	\begin{equation}
	\begin{split}
	& (\hat{\theta}_{f},\hat{\theta}_{c},\hat{\theta}_{y},\hat{\theta}_u)= \arg\max_{\theta_f,\theta_c,\theta_y,\theta_u}(C(\theta_f,\theta_c,\theta_y,\theta_u))\\
	\end{split}
	\end{equation}
	%%%%%%%%%%%%%%%%%%%%%%%%%%%%%%%%%%%%%%%%%%%%%%%%%%%%%%%%%%%%%%%%%%%%%%%%%%%%%%%%%%%%
	% \begin{figure}[ht]
	%  \small
	%  \centering
	%  \begin{tabular}[b]{ c}
	% %  (a) MC-BMN \\
	% \includegraphics[width=0.4\textwidth]{fig/certainDistanceOTHER.png}\\
	% %  (b) MC-BMN-2 \\ 
	% \vspace{-2em}
	% \includegraphics[width=0.4\textwidth]{fig/certainDistanceYESNO.png}
	%   \end{tabular}
	% \caption{Upper Graph: Certainty Improvement for OTHER type and Lower graph:  Certainty Improvement for Yes/No type}
	% % 	  \vspace{-0.7cm}
	%   \label{fig:my_label}
	% %   \vspace{-2em}
	% \end{figure}
	
	% \begin{figure}
	%     \centering
	%     \includegraphics[width=0.45\textwidth]{latex/fig/certainDistanceOTHER.png}
	%     \caption{Certainty Improvement for OTHER type}
	%     \label{fig:my_label}
	% \end{figure}
	
	% \begin{figure}
	%     \centering
	%     \includegraphics[width=0.45\textwidth]{latex/fig/certainDistanceYESNO.png}
	%     \caption{Certainty Improvement for Yes/No type}
	%     \label{fig:my_label1}
	% \end{figure}
	%%%%%%%%%%%%%%%%%%%%%%%%%%%%%%%%%%%%%%%%%%%%%%%%%%%%%%%%%%%%%%%%%%%%%%%%%%%%%%%%%%%%
	%%%%%%%%%%%%%%%%%%%%%%%%%%%%%%%%%%%%%%%%%%%%%%%%%%%%%%%%%%%%%%%%%%%%%%%%%%%%%%%%%%%%
	\vspace{-1em}
	\section{Experiments}
\begin{figure*}[ht]
		\small
		\centering
		\begin{tabular}[b]{ c  c  c c}
			(a) Classification error & (b) Misclassified & (c) CD-Others & (d) CD-Yes/No\\ 
			\includegraphics[width=0.24\textwidth, height=0.15\textheight]{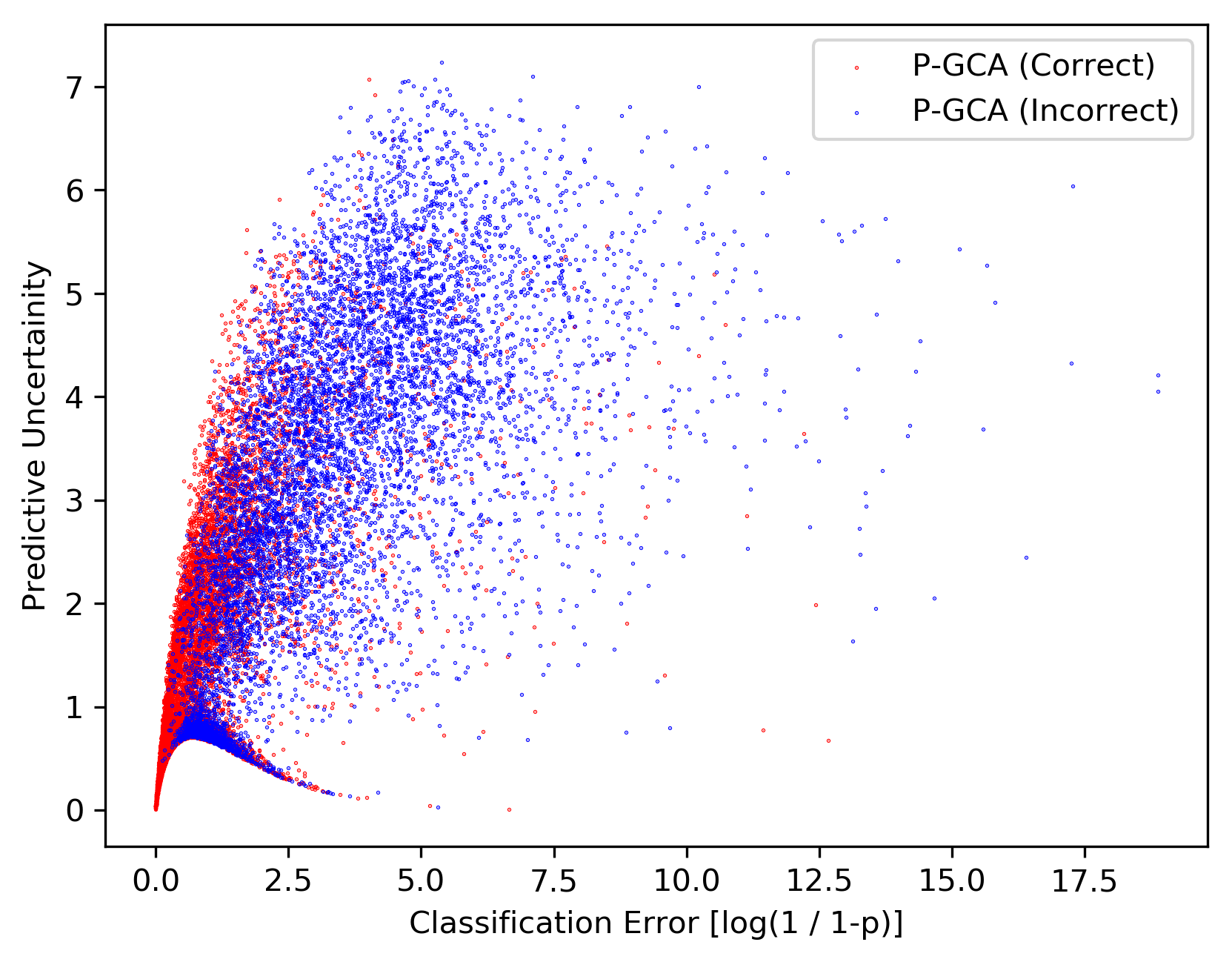}
			& \includegraphics[width=0.24\textwidth, height=0.15\textheight]{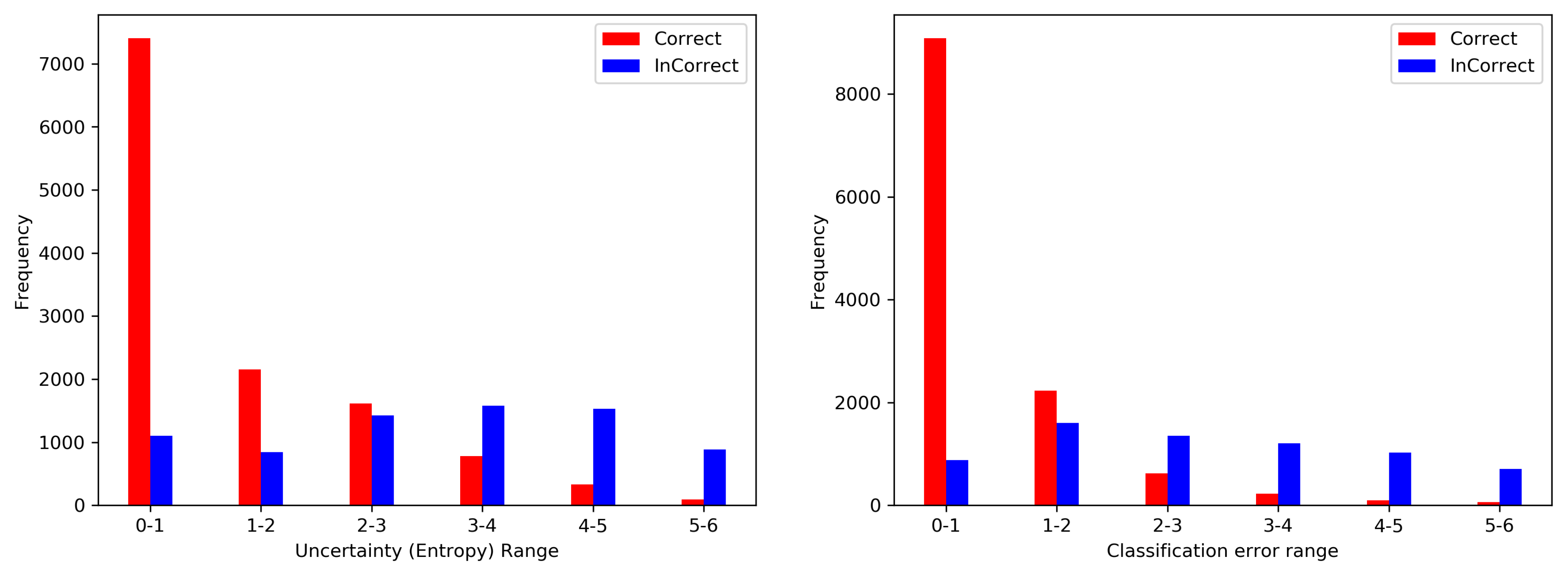}
			& \includegraphics[width=0.24\textwidth, height=0.15\textheight]{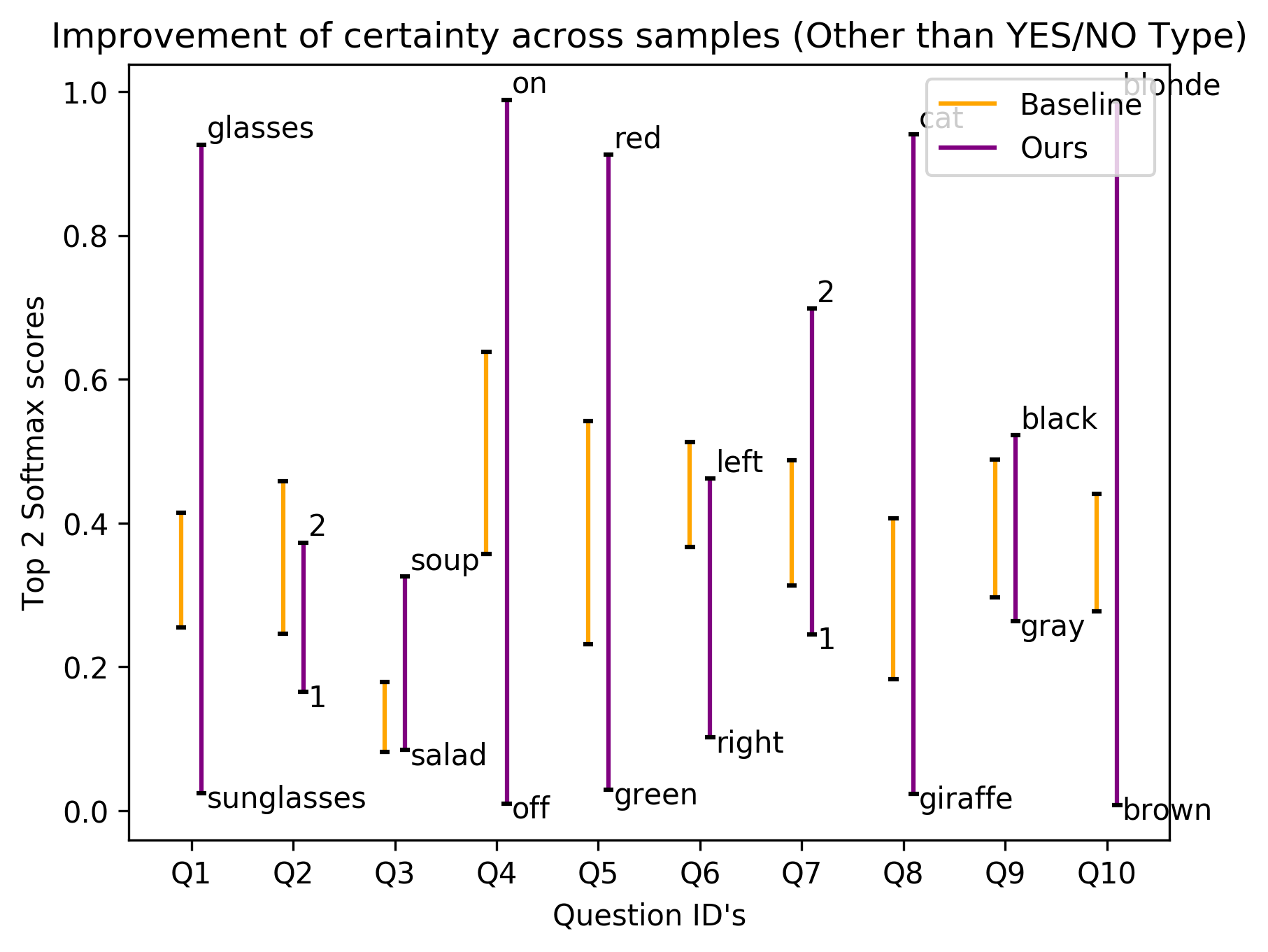}
			& \includegraphics[width=0.24\textwidth, height=0.15\textheight]{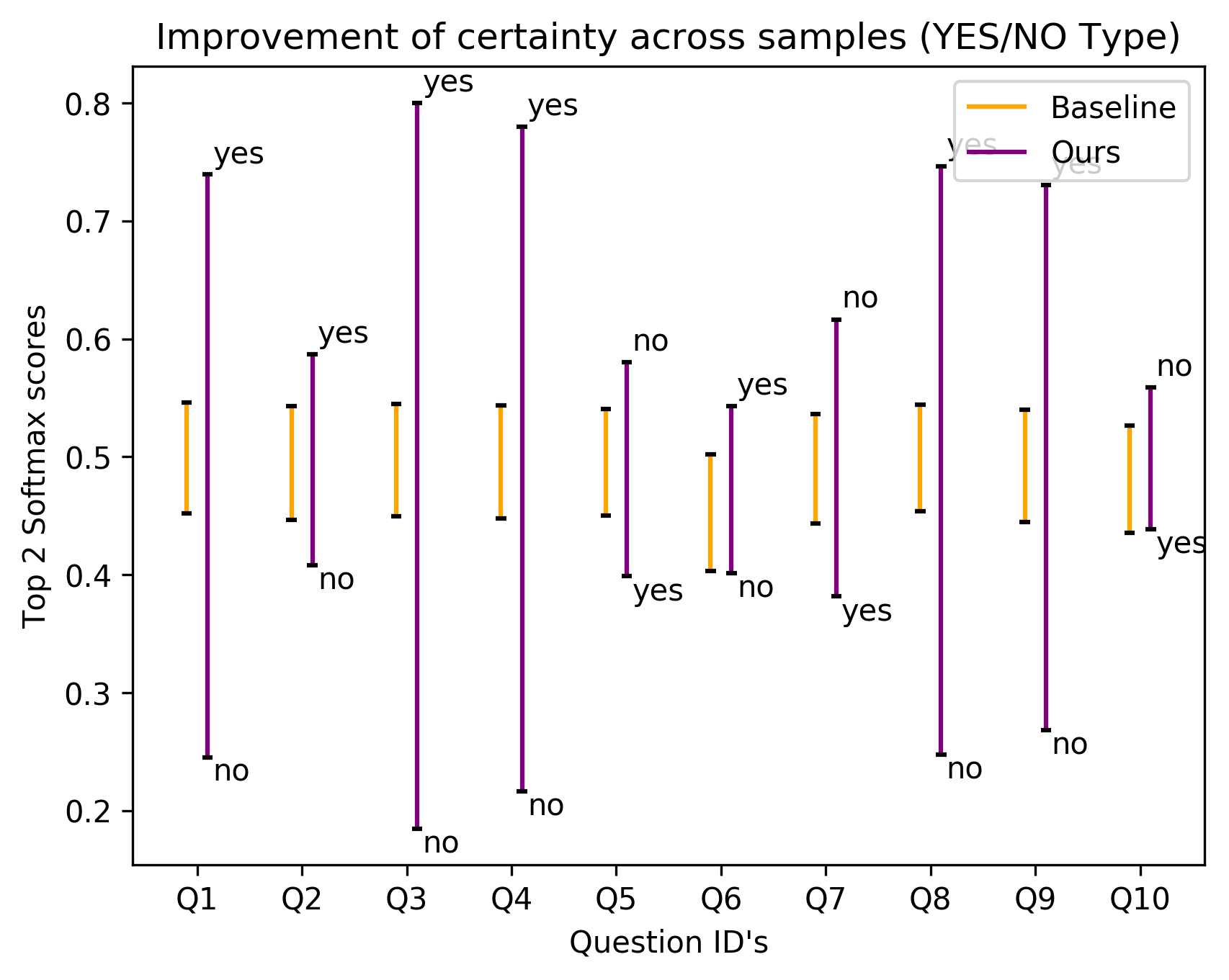}
		\end{tabular}
		\caption{(a) Uncertainty vs Classification Error plots for our network for 20,000 randomly sampled images. We drew 25 samples of each image using Monte-Carlo sampling from the distribution. (b) Plots showing frequency of samples vs Uncertainty and frequency of samples vs Classification error respectively (c) Distance between the Top 2 Softmax scores for some Questions of type other than \textit{yes/no} (d) Distance between the Top 2 Softmax scores for some Questions of type \textit{yes/no} (Questions corresponding to (c) and (d) could be found in supplementary.) } 
		\label{tbl:variance}
		\vspace{-1.5em}
	\end{figure*}

	% We evaluate the proposed method GCA in some ways which include both quantitative analysis and qualitative analysis.
	We evaluate the proposed GCA methods and have provided both quantitative analysis and qualitative analysis. 
	The former includes: i) Ablation analysis of proposed models (Section-~\ref{variants}), ii) Analysis of uncertainty effect on answer predictions (Figure-~\ref{tbl:variance} (a,b)), iii) Differences of Top-2 softmax scores for answers for some representative questions (Figure-~\ref{tbl:variance} (c,d)) and iv) Comparison of attention map of our proposed uncertainty model against other variants using Rank correlation (RC) and  Earth Mover Distance (EMD) ~\cite{arjovsky_STAT2017wasserstein} as shown in Table-\ref{tab_rank_correlation} for VQA-HAT~\cite{Das_EMNLP2016} and in Table-~\ref{tab_vqa-x_rc} for VQA-X~\cite{huk2018multimodal}  . Finally, we compare PGCA with state of the art methods as mentioned in Section-\ref{SOTA} . Qualitative analysis includes visualization of certainty activation maps for some representative images as we move from our basic model to the P-GCA model. (Section \ref{ResultQual})
	%%%%%%%%%%%%%%%%%%%%%%%%%%%%%%%%%%%%%%%%%%%%%%%%%%%%%%%%%%%%%%%%%%%%%%%%%%%%%%%%%%%%%%%%%%%%%%%%%%%%%%%%	
	\subsection{Datasets}
	\textbf{VQA-v1~\cite{VQA}}: We conduct our experiments on VQA benchmark VQA-v1~\cite{VQA} dataset, which contains human-annotated questions and answers based on images on MS-COCO dataset. This dataset includes 2,04,721 images in total, out of which 82,783 images are for training, 40,504 images for validation, and 81,434 images for testing. Each image is associated with three questions, and each question has ten possible answers. There are 248349 Question-Answer pairs for training, 121512 pairs for validation, and 244302 pairs for testing. 
	
	\textbf{VQA-v2~\cite{Goyal_CVPR2017}}: We provide benchmark result on VQA-v2~\cite{Goyal_CVPR2017} dataset. This dataset removes bias present in VQA-v1 by adding a conjugate image pair. It contains 443,757 image-question pairs on the training set, 214,354 pairs on the validation set and 447,793 pairs on the test set, which is more than twice the first version. All the questions and answers pairs are annotated by human annotators. The benchmark results on VQA-v2 dataset is presented in Table-\ref{VQA2_accuracy}.
	
	\textbf{VQA-HAT~\cite{Das_EMNLP2016}}: To compare our attention map with human-annotated attention maps, we use VQA-HAT~\cite{Das_EMNLP2016} dataset. This dataset is developed for image de-blurring for answering the visual question. It contains 58475 human-annotated attention maps out of  248349 training examples and includes three sets of 1374 human-annotated attention maps out of 121512 validation examples of question image pairs in the validation dataset. This dataset is developed for VQA-v1 only. 
	%%%%%%%%%%%%%%%%%%%%%%%%%%%%%%%%%%%%%%%%%%%%%%%%%%%%%%%%%%%%%%%%%%%%%%%%%%%%%%%%%%%%%%%%%%%%%%%%%%%%%%%%	
	
	%\vspace{-1em}
	\begin{table}[ht]
		\begin{center}			
			\begin{tabular}{|l|c|c|c|c|  } \hline
				\textbf{Models} & \textbf{All} & \textbf{Yes/No} & \textbf{Number} & \textbf{Others}  \\ \hline 
				Baseline & {63.8}& { 82.2 }& { 37.3 }& { 54.2 } \\
				VE & { 64.1 }& {82.3 }& { 37.2}& { 54.3} \\
				UDL & { 64.4 }& {82.6 }& { 37.2}& { 54.5} \\
				AUL   &{ 64.7 }& {82.9 }& { 37.4 }& { 54.6} \\
				PUL  & { 64.9 }& {83.0 }& { 37.5 }& { 54.6} \\ \hline
				UDL+VE	 & 64.8 & 82.8 & 37.4& 54.5 \\
				AUL+VE  & 65.0 & 83.3 & 37.8 & 54.7 \\
				PUL+ VE  & 65.3 & 83.3 & 37.9 & 54.9 \\ 
				AUL +UDL & 65.6 & 83.3 & 37.6 &  55.0 \\ 
				PUL + UDL  & 65.9 & 83.7 & 37.8 & 55.2 \\ \hline
				A-GCA (ours)& 66.3 & 84.2 & 38.0 &  55.5 \\ 
				P-GCA (ours)& \textbf{66.5} & \textbf{84.7} &\textbf{ 38.4} &  \textbf{55.9}\\\hline
			\end{tabular}
			% 		\vspace{-1em}
			% \vspace{0.3cm}
			\caption{Ablation analysis for Open-Ended VQA1.0 accuracy on test-dev}
			\label{abl_VQA1_accuracy}
			\vspace{-1.75em}
		\end{center}		
	\end{table}
	\begin{table}
		\centering
		\begin{tabular}{|l|c|c|}
			\hline
			\textbf{Model} & \textbf{RC($\uparrow$)} & \textbf{EMD($\downarrow$)}\\
			\hline
			Baseline  & 0.3017 & 0.3825 \\
			Deconv ReLU  & 0.3198 & 0.3801 \\
			Guided GradCAM& 0.3275 & 0.3781\\
			Aleatoric mask & 0.3571 & 0.3763 \\
			Predictive mask  & \textbf{0.3718} & \textbf{0.3714} \\\hline
		\end{tabular}
		%   \vspace{-0.3cm}
		\caption{Rank Correlation for explanation mask in VQA-X ~\cite{huk2018multimodal} data with our explanation mask using Grad-Cam.}
		\label{tab_vqa-x_rc}
		\vspace{-1.52em}
	\end{table}

	\subsection{Ablation Analysis for Uncertainty }\label{variants}
	%%%%%%%%%%%%%%%%%%%%%%%%%%%%%%%%%%%%%%%%%%%%%%%%%%%%%%%%%%%%%%%%%%%%%%%%%%%%%%%%%%%%%%%%%%%%%%%%%%%%%%
	Our proposed GCA model's loss consists of undistorted and distorted loss. The undistorted loss is the Standard Cross-Entropy (SCE) loss. The distorted loss consists of uncertain loss (either aleatoric uncertainty loss (AUL), or predictive uncertainly loss (PUL)), Variance Equalizer (VE) loss and Uncertainty Distorted loss (UDL).  In the first block of the Table-~\ref{abl_VQA1_accuracy}, we report the results when these losses are used individually. (Only SCE loss is there in the Baseline). We use a variant of the MCB \cite{Fukui_arXiv2016} model as our baseline method. As seen, PUL, when used individually, outperforms the other 4. This could be attributed to PUL guiding the model to minimize both the data and the model uncertainty. The second block of the Table-~\ref{abl_VQA1_accuracy} depicts the results when we tried while combining two different individual losses. The model variant, which is guided using the combination of PUL and UDL loss performs best among the five variants. Then finally, after combining (AUL+UDL+VE+SCE), denoting it as A-GCA model and combining (PUL+UDL+VE+SCE), indicating it as P-GCA, we report an improvement of around 2.5\% and 2.7\% accuracy score respectively.
	
	% The behavior of individual loss with respect to uncertainty and the ablation analysis of epistemic and aleatoric uncertainty comparison, analysis of noise on both uncertainty is provided in the supplementary material. 
	%%%%%%%%%%%%%%%%%%%%%%%%%%%%%%%%%%%%%%%%%%%%%%%%%%%%%%%%%%%%%%%%%%%%%%%%%%%%%
	\begin{table}
		\centering
		\begin{tabular}{|l|c|c|c|}
			\hline
			\textbf{Model} & \textbf{RC($\uparrow$)} & \textbf{EMD($\downarrow$) }& \textbf{CD($\uparrow$)}\\
			\hline
			SAN \cite{Das_EMNLP2016}& 0.2432 & 0.4013 &--\\
			CoAtt-W\cite{Lu_NIPS2016}	& {0.246 } & --& --\\		
			CoAtt-P \cite{Lu_NIPS2016}	& {0.256 }&--& --\\
			CoAtt-Q\cite {Lu_NIPS2016}	& {0.264 }&--&--\\
			DVQA(K=1)\cite{Patro_CVPR2018dvqa}& {0.328}&--&--\\\hline
			Baseline (MCB) & 0.2790 & 0.3931& -- \\
			VE (ours) & 0.2832 & 0.3931& 0.1013 \\
			UDL (ours) & 0.2850 & 0.3914& 0.1229 \\
			AUL (ours) & 0.2937 & 0.3867 & 0.1502\\
			PUL(ours) & 0.3012 & 0.3805& 0.1585 \\
			PUL + VE (ours) & 0.3139 & 0.3851 & 0.1631\\
			PUL + UDL(ours)& 0.3243 & 0.3824 & 0.1630\\\hline
			A-GCA (ours)& 0.3311 & 0.3784 & 0.1683\\
			P-GCA (ours) & \textbf{0.3341} & \textbf{0.3721} & \textbf{0.1710}\\\hline
			Human \cite{Das_EMNLP2016} & { 0.623 } &--& --\\ \hline
		\end{tabular}
		%   \vspace{0.3cm}
		\caption{Ablation analysis and SOTA between HAT\cite{Das_EMNLP2016} attention and generated attention mask}
		\label{tab_rank_correlation}
		\vspace{-1.5em}
	\end{table}
	Further, we plotted Predictive uncertainty (Figure-~\ref{tbl:variance}(a,b)) of some randomly chosen samples against the Classification error (error=$ \log{\frac{1}{1-p}}$, where \textit{p} is the probability of misclassification). As seen, when the samples are correct, they are also certain and have less Classification Error (CE). To visualize the direct effect of decreased uncertainty, we plotted (Figure-~\ref{tbl:variance}(c, d)). It can be seen that how similar classes like (glasses, sunglasses) and (black, gray), etc., thus leading to uncertainty, got separated more in the logit space in the proposed model.

	\subsection{Analysis of Attention Maps}\label{AnalysisQuant}
	We compare attention maps produced by our proposed GCA model, and it's variants with the base model and reports them in Table-\ref{tab_rank_correlation}. Rank correlation and EMD score are calculated for the produced attention map against human-annotated attention (HAT) maps ~\cite{Das_EMNLP2016}. In the table, as we approach the best-proposed GCA model, Rank correlation (RC) is increasing. EMD is also decreasing (Lower the better) as we move towards GCA. To verify our intuition, that we can learn better attention mask by minimizing the uncertainty present in the attention mask, we start with VE and observe that both rank correlation and answer accuracy increase by 0.42 and 0.3 \% from baseline respectively. We also observe that with UDL, AUL, and PUL based loss minimization technique, both RC and EMD improves, as shown in the Table-~\ref{tab_rank_correlation}.
	% Thus, the proposed Aleatoric-GCA method is improving the attention globally. % concerning Grad-CAM.
	Aleatoric-GCA (A-GCA) improves 5.21\% in terms of RC and 2.5\% in terms of accuracy.  Finally, the proposed Predictive-GCA (P-GCA), which is modeled to consider both data and the model uncertainty improves the RC by 5.51\% and accuracy by 2.7\% as shown in the Table-~\ref{tab_rank_correlation} and Table-~\ref{abl_VQA1_accuracy}. Since HAT maps are only available for VQA-v1 dataset, thus, this ablation analysis has been performed only for VQA-v1. We also providing SOTA results for VQA-v1 and VQA-v2 dataset as shown in Table-~\ref{VQA1_accuracy} and Table-~\ref{VQA2_accuracy} respectively. 
	Also, we compare with our gradient certainty explanation with human explanation present in VQA-v2 dataset for the various model as mentioned in Table-~\ref{tab_vqa-x_rc}. This human explanation mask only available for VQA-v2 dataset. We observe that our attention (P-GCA) mask performs better than others as well.  The evaluation methods for VQA dataset and HAT dataset are provided in supplementary material.

	\begin{table}
		\begin{center}			
			\begin{tabular}{|l|c|c|c|c|} \hline
				\textbf{Models} & \textbf{All} & \textbf{Y/N} & \textbf{Num} & \textbf{Oth}  \\ \hline 
				%Baseline-ATT & { 56.7 }& {78.9 }& { 35.2}& { 36.4} \\
				%   LSTM Q+I+ Attention(LQIA) & { 56.1 }& { 80.3 }& { 37.4 }& {  40.4} \\
				DPPnet \cite{Noh_CVPR2016}    & { 57.2 }& {80.7 }& { 37.2 }& { 41.7} \\
				SMem[\cite{Xu_ECCV2016}]      & { 58.0 }& {80.9 }& { 37.3 }& { 43.1} \\ 
				SAN \cite{Yang_CVPR2016}      & { 58.7 }& {79.3 }& { 36.6 }& { 46.1 }\\
				% 			QRU(1)\cite{Li_NIPS2016}  & { 59.3 }& {81.0 }& { 35.9 }& { 46.0 }\\
				% DAN(K=4)+ LQIA & \textbf{60.2 }& { 80.9 }& \textbf{ 37.4 }& \textbf{ 47.2 } \\ \hline
				DMN\cite{Xiong_arXiv2016} & {60.3 }& { 80.5 }& { 36.8 }& { 48.3 } \\ 
				QRU(2)\cite{Li_NIPS2016}  & { 60.7 }& {82.3 }& { 37.0 }& { 47.7 }\\				
				% DCN Mul\_v2(K=4)+LQIA  & \textbf{60.9 }& {81.3 }& \textbf{ 37.5 }& \textbf{ 48.2 } \\ \hline
				HieCoAtt \cite{Lu_NIPS2016} & {61.8}& { 79.7 }& { 38.9 }& { 51.7 } \\
				MCB \cite{Fukui_arXiv2016} & {64.2}& { 82.2 }& { 37.7 }& { 54.8 } \\
				MLB \cite{Kim_ICLR2017} & {65.0}& { 84.0} & { 37.9 }& { 54.7 }\\
				DVQA\cite{Patro_CVPR2018dvqa}&65.4& 83.8& 38.1& 55.2
				\\ \hline
				% 			AAN + SAN (ours)& 62.3 & 80.4 & 37.2 &  49.8 \\ 
				P-GCA + SAN (ours)& 60.4 & 80.7 & 36.6 & 47.9 \\ 
				A-GCA + MCB (ours)& 66.3 & 84.2 & 38.0 &  55.5 \\
				P-GCA + MCB (ours)& \textbf{66.5} & \textbf{84.6} &\textbf{ 38.4} &  \textbf{55.9} \\ \hline
			\end{tabular}
			% 		\vspace{-1em}
			\caption{SOTA: Open-Ended VQA1.0 accuracy on test-dev}
			\label{VQA1_accuracy}
			\vspace{-2.42em}
		\end{center}		
	\end{table}
	% 	\vspace{-1em}
	%%%%%%%%%%%%%%%%%%%%%%%%%%%%%%%%%%%%%%%%%%%%%%%%%%%%%%%%%%%%%%%%%%%%%%%%%%%%%%%%%%%%%%%%%%%%%%%%%%%%%%
	\subsection{Comparison with baseline and state-of-the-art}\label{SOTA}
	%%%%%%%%%%%%%%%%%%%%%%%%%%%%%%%%%%%%%%%%%%%%%%%%%%%%%%%%%%%%%%%%%%%%%%%%%%%%%%%%%%%%%%%%%%%%%%%%%%%%%%
	% We provided comparison of our proposed models with the state of the art models in table \ref{VQA1_accuracy}. Last 2 rows in the table contains results of our models and rest are state of the art models. We observe an improvement of $\sim7\%$ from the BaseLine-ATT network and $\sim5\%$ from SAN. Here, BaseLine-ATT is our implementation of the SAN with only one attention layer.
	We obtain the initial comparison with the baselines on the rank correlation on human attention (HAT) dataset~\cite{Das_EMNLP2016} that provides human attention while solving for VQA. Between humans, the rank correlation is 62.3\%. The comparison of various state-of-the-art methods and baselines are provided in Table~\ref{tab_rank_correlation}. We use a variant of MCB~\cite{Fukui_arXiv2016} model as our baseline method. We obtain an improvement of around 5.2\% using A-GCA model and 5.51\% using P-GCA model in terms of rank correlation with human attention. From this, we justify that our attention map is more similar to human attention map. We also compare with the baselines on the answer accuracy on VQA-v1\cite{VQA} dataset, as shown in Table-~\ref{VQA1_accuracy}. We obtain an improvement of around 2.7\% over the comparable MCB baseline. Our MCB based model A-GCA and P-GCA improves by 0.9\% and 1.1\% accuracy as compared to state of the art model DVQA~\cite{Patro_CVPR2018dvqa} on VQA-v1. However, using a saliency-based method  ~\cite{Judd_ICCV2009} that is trained on eye-tracking data to obtain a measure of where people look in a task-independent manner, results in more correlation with human attention (0.49), as noted by \cite{Das_EMNLP2016}. However, this is explicitly trained using human attention and is not task-dependent. In our approach, we aim to obtain a method that can simulate human cognitive abilities for solving the tasks. We provide state of the art results for VQA-v2 in Table-~\ref{VQA2_accuracy}. This table shows that using GCA method, the VQA result improves. We have provided more results for attention map visualization for both types of uncertainty methods here\footnote{\url{https://delta-lab-iitk.github.io/U-CAM/} }.
	\begin{table}
		\begin{center}			
			\begin{tabular}{|l|c|c|c|c|  } \hline
				\textbf{Models} & \textbf{All} & \textbf{Y/N} & \textbf{Num} & \textbf{Oth}  \\ \hline 
				SAN-2\cite{Yang_CVPR2016}     & {56.9 }& {74.1 }& {35.5 }& {44.5 }\\ 
				MCB  \cite{Fukui_arXiv2016} & {64.0}& { 78.8 }& {38.3 }& {53.3 } \\
				Bottom[\cite{anderson_CVPR2018bottom}]&65.3 & 81.8& 44.2& 56.0\\
				DVQA\cite{Patro_CVPR2018dvqa}&65.9& 82.4& 43.2& 56.8\\
				% 			MUTAN[\citeauthor{Ben_ICCV2017}] & {66.0}& { 82.8 }& {44.5}& { 56.5 } \\
				MLB \cite{Kim_ICLR2017} & {66.3}& {83.6} & {44.9}& {56.3 } \\
				DA-NTN \cite{bai_ECCV2018deep} & {67.5}& {84.3} & {47.1}& {57.9} \\
				Counter\cite{zhang_ICLR2018learning}& 68.0&83.1&51.6&58.9\\
				BAN\cite{kim_NIPS2018bilinear} &\textbf{ 69.5}& 85.3&\textbf{50.9}&\textbf{60.2}\\\hline
				P-GCA + SAN (ours)& 59.2 &75.7 & 36.6 & 46.8 \\ 
				P-GCA + MCB (ours) & 65.7 & 79.6 & 40.1 &  54.7 \\ 
				P-GCA + Counter (ours)& 69.2 & \textbf{85.4} & 50.1 & 59.4 \\\hline
			\end{tabular}
			% 		\vspace{-1em}
			\caption{SOTA: Open-Ended VQA2.0 accuracy on test-dev}
			\label{VQA2_accuracy}
			\vspace{-2.42em}
			% 		\vspace{-2em}
		\end{center}		
	\end{table}
	\vspace{-.5em}
	%%%%%%%%%%%%%%%%%%%%%%%%%%%%%%%%%%%%%%%%%%%%%%%%%%%%%%%%%%%%%%%%%%%%%%% 
	\subsection{ Training  and Model Configuration}
	\vspace{-.3em}
	We trained the  P-GCA model using classification loss and uncertainty loss in an end-to-end manner. We have used ADAM  optimizer to update the classification model parameter and configured hyper-parameter values using validation dataset as follows: \{learning rate = 0.0001, batch size = 200, beta = 0.95, alpha = 0.99 and epsilon = 1e-8\} to train the classification model. We have used SGD  optimizer to update the uncertainty model parameter and configured hyper-parameter values using validation dataset as follows: \{learning rate = 0.004, batch size = 200, and epsilon = 1e-8\} to train the uncertainty model.
	
	%%%%%%%%%%%%%%%%%%%%%%%%%%%%%%%%%%%%%%%%%%%%%%%%%%%%%%%%%%%%%%%%%%%%%%%%%%%%%%%%%%%%%%%%%%%%%%%%%%%%%%%%	
	\begin{figure}[!htb]
		%\vspace{1in}
		\centering
		\includegraphics[width=0.45\textwidth]{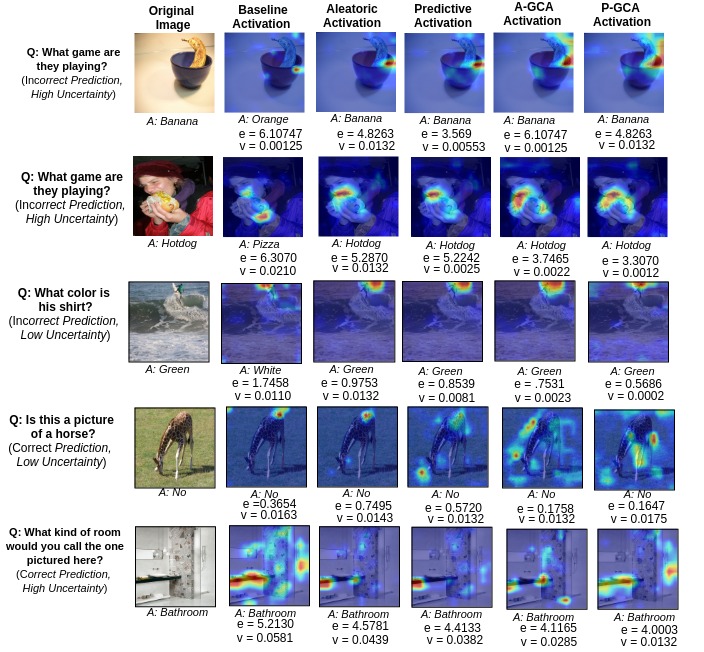}
		% 	\caption{Examples with different approaches in each column for improving attention using explanation in a self supervised manner. The first column indicates the given target image and its question and answer. Starting from second column, it indicates the attention map for Stack Attention Network, MSE based approach, Coral based approach, MMD based approach, Adversarial based approach respectively.}
		% 	\label{fig:result_1_A}
		% \end{figure*}
		% \begin{figure*}[!htb]
		%\vspace{1in}
		% 	\centering
		% 		\vspace{-0.5em}
		% {-----------------------------------------------------------------}
		% 		\vspace{-1em}
		% 	\includegraphics[width=0.45\textwidth]{fig/all_grad_visual_1_5_9.png}
		% 	\vspace{-1.2em}
		\caption{Examples with different approaches in each column for improving attention using explanation in a self-supervised manner. The first column indicates the given target image and its question and answer. Starting from the second column, it shows the activation map for baseline (MCB) Attention Network, Aleatoric (AUL), Predictive (PUL), A-GCA, P-GCA based approach respectively.}
		\label{fig:result_1_B}
		\vspace{-1.2em}
	\end{figure}
	
	%%%%%%%%%%%%%%%%%%%%%%%%%%%%%%%%%%%%%%%%%%%%%%%%%%%%%%%%%%%%%%%%%%%%%%%%%%
	\subsection{Qualitative Result}\label{ResultQual}
	%%%%%%%%%%%%%%%%%%%%%%%%%%%%%%%%%%%%%%%%%%%%%%%%%%%%%%%%%%%%%%%%%%%%%%%%%%
	We provide attention map visualization of all models for 5 example images, as shown in Figure-~\ref{fig:result_1_B}. The first raw, the baseline model misclassifies the answer due to high uncertainty value, that gets resolved by our methods(P-GCA). We can see how attention is improved as we go from our baseline model (MCB) to the proposed Gradient Certainty model (P-GCA). For example, in the first row, MCB is unable to focus on any specific portion of the image, but as we go towards the right, it focuses the cup bottom, (indicated by intense orange color in the map). Same can be seen for other images also. We have visualized Grad-CAM maps to support our hypothesis that Grad-CAM is a very good way for visualizing what the network learns as it can focus on right portions of the image even in the baseline model (MCB), and therefore, can be used as a tutor to improve attention maps. For example, in MCB it tries to focus on the right portions but with the focus to other points as well. However, in our proposed model, visualization improves as the models focuses only on the required portion.
	\vspace{-1.2em}
	\section{Conclusion}
	\vspace{-.5em}
	%%%%%%%%%%%%%%%%%%%%%%%%%%%%%%%%%%%%%%%%%%%%%%%%%%%%%%%%%%%%%%%%%%%%%%%%%%
	In this paper, we provide a method that uses gradient-based certainty attention regions to obtain improved visual question answering. The proposed method yields improved uncertainty estimates that are correspondingly more certain or uncertain, show consistent correlation with misclassification and are focused quantitatively on better attention regions as compared to other states of the art methods. The proposed architecture can be easily incorporated in various existing VQA methods as we show by incorporating the method in SAN \cite{Yang_CVPR2016} and MCB \cite{Fukui_arXiv2016} models. The proposed technique could be used as a general means for obtaining improved uncertainty and explanation regions for various vision and language tasks, and in future, we aim to evaluate this further for other tasks such as `Visual Dialog' and image captioning tasks.
	% In this paper we have proposed a method to obtain surrogate supervision for obtaining improved attention using visual explanation. The proposed method shows that the improved attention indeed results in improved results for the semantic task such as VQA or Visual dialog. Our method provides an initial means for obtaining surrogate supervision for attention and in future we would like to further investigate other means of obtaining improved attention.
	%%%%%%%%%%%%%%%%%%%%%%%%%%%%%%%%%%%%%%%%%%%%%%%%%%%%%%%%%%%%%%%%%%%%%%%%%%
	%%%%%%%%%%%%%%%%%%%%%%%%%%%%%%%%%%%%%%%%%%%%%%%%%%%%%%%%%%	
	\vspace{-0.5cm}
	\section{Acknowledgment}
	\vspace{-0.3cm}
	We acknowledge the help provided by Delta Lab members and our family who have supported us in this research activity. 
%%%%%%%%%%%%%%%%%%%%%%%%%%%%%%%%%%%%%%%%%%%%%%%%%%%%%%%%%%%%%%%%%%%%%%%%%%
{\small
	\bibliographystyle{ieee_fullname}
	\bibliography{egbib}

\begin{thebibliography}{10}\itemsep=-1pt

\bibitem{anderson_CVPR2018bottom}
Peter Anderson, Xiaodong He, Chris Buehler, Damien Teney, Mark Johnson, Stephen
  Gould, and Lei Zhang.
\newblock Bottom-up and top-down attention for image captioning and visual
  question answering.
\newblock In {\em Proceedings of the IEEE Conference on Computer Vision and
  Pattern Recognition}, pages 6077--6086, 2018.

\bibitem{VQA}
Stanislaw Antol, Aishwarya Agrawal, Jiasen Lu, Margaret Mitchell, Dhruv Batra,
  C.~Lawrence Zitnick, and Devi Parikh.
\newblock {VQA}: {V}isual {Q}uestion {A}nswering.
\newblock In {\em International Conference on Computer Vision (ICCV)}, 2015.

\bibitem{arjovsky_STAT2017wasserstein}
Martin Arjovsky, Soumith Chintala, and L{\'e}on Bottou.
\newblock Wasserstein gan.
\newblock {\em stat}, 1050:26, 2017.

\bibitem{bai_ECCV2018deep}
Yalong Bai, Jianlong Fu, Tiejun Zhao, and Tao Mei.
\newblock Deep attention neural tensor network for visual question answering.
\newblock In {\em Proceedings of the European Conference on Computer Vision
  (ECCV)}, pages 20--35, 2018.

\bibitem{Barnard_JMLR2003}
K Barnard, P Duygulu, and D Forsyth.
\newblock N. de freitas, d.
\newblock {\em Blei, and MI Jordan," Matching Words and Pictures", submitted to
  JMLR}, 2003.

\bibitem{blundell_ICML2015weight}
Charles Blundell, Julien Cornebise, Koray Kavukcuoglu, and Daan Wierstra.
\newblock Weight uncertainty in neural network.
\newblock In {\em International Conference on Machine Learning}, pages
  1613--1622, 2015.

\bibitem{Chen_CVPR2015}
Xinlei Chen and C Lawrence~Zitnick.
\newblock Mind's eye: A recurrent visual representation for image caption
  generation.
\newblock In {\em Proceedings of the IEEE conference on computer vision and
  pattern recognition}, pages 2422--2431, 2015.

\bibitem{Das_EMNLP2016}
Abhishek Das, Harsh Agrawal, C.~Lawrence Zitnick, Devi Parikh, and Dhruv Batra.
\newblock {Human Attention in Visual Question Answering: Do Humans and Deep
  Networks Look at the Same Regions?}
\newblock In {\em Conference on Empirical Methods in Natural Language
  Processing (EMNLP)}, 2016.

\bibitem{Das_ICCV2017}
Abhishek Das, Satwik Kottur, Jos\'e~M.F. Moura, Stefan Lee, and Dhruv Batra.
\newblock Learning cooperative visual dialog agents with deep reinforcement
  learning.
\newblock In {\em IEEE International Conference on Computer Vision (ICCV)},
  2017.

\bibitem{Dorman_github2018}
Kyle Dorman.
\newblock Bayesian neural network blogpost.
\newblock {\em
  https://github.com/kyle-dorman/bayesian-neural-network-blogpost}.
\newblock Accessed: 2018-07-015.

\bibitem{Fang_CVPR2015}
Hao Fang, Saurabh Gupta, Forrest Iandola, Rupesh Srivastava, Li Deng, Piotr
  Doll{\'a}r, Jianfeng Gao, Xiaodong He, Margaret Mitchell, John Platt, et~al.
\newblock From captions to visual concepts and back.
\newblock In {\em Proceedings of the IEEE conference on computer vision and
  pattern recognition}, 2015.

\bibitem{Farhadi_ECCV2010}
Ali Farhadi, Mohsen Hejrati, Mohammad~Amin Sadeghi, Peter Young, Cyrus
  Rashtchian, Julia Hockenmaier, and David Forsyth.
\newblock Every picture tells a story: Generating sentences from images.
\newblock In {\em European conference on computer vision}, pages 15--29.
  Springer, 2010.

\bibitem{Fortunato_Arxiv2017}
Meire Fortunato, Charles Blundell, and Oriol Vinyals.
\newblock Bayesian recurrent neural networks.
\newblock {\em arXiv preprint arXiv:1704.02798}, 2017.

\bibitem{Fukui_arXiv2016}
Akira Fukui, Dong~Huk Park, Daylen Yang, Anna Rohrbach, Trevor Darrell, and
  Marcus Rohrbach.
\newblock Multimodal compact bilinear pooling for visual question answering and
  visual grounding.
\newblock {\em arXiv preprint arXiv:1606.01847}, 2016.

\bibitem{Gal2016Uncertainty}
Yarin Gal.
\newblock {\em Uncertainty in Deep Learning}.
\newblock PhD thesis, University of Cambridge, 2016.

\bibitem{Gal_ICML2016}
Yarin Gal and Zoubin Ghahramani.
\newblock Dropout as a bayesian approximation: Representing model uncertainty
  in deep learning.
\newblock In {\em International Conference on Machine Learning (ICML)}, pages
  1050--1059, 2016.

\bibitem{Gal_NIPS2016}
Yarin Gal and Zoubin Ghahramani.
\newblock A theoretically grounded application of dropout in recurrent neural
  networks.
\newblock In {\em Advances in neural information processing systems}, pages
  1019--1027, 2016.

\bibitem{Goyal_CVPR2017}
Yash Goyal, Tejas Khot, Douglas Summers-Stay, Dhruv Batra, and Devi Parikh.
\newblock Making the v in vqa matter: Elevating the role of image understanding
  in visual question answering.
\newblock In {\em Proceedings of the IEEE Conference on Computer Vision and
  Pattern Recognition}, pages 1--9, 2017.

\bibitem{huk2018multimodal}
Dong Huk~Park, Lisa Anne~Hendricks, Zeynep Akata, Anna Rohrbach, Bernt Schiele,
  Trevor Darrell, and Marcus Rohrbach.
\newblock Multimodal explanations: Justifying decisions and pointing to the
  evidence.
\newblock In {\em Proceedings of the IEEE Conference on Computer Vision and
  Pattern Recognition}, pages 8779--8788, 2018.

\bibitem{Johnson_CVPR2016}
Justin Johnson, Andrej Karpathy, and Li Fei-Fei.
\newblock Densecap: Fully convolutional localization networks for dense
  captioning.
\newblock In {\em Proceedings of the IEEE Conference on Computer Vision and
  Pattern Recognition}, pages 4565--4574, 2016.

\bibitem{Judd_ICCV2009}
Tilke Judd, Krista Ehinger, Fr{\'e}do Durand, and Antonio Torralba.
\newblock Learning to predict where humans look.
\newblock In {\em Computer Vision, 2009 IEEE 12th international conference on},
  pages 2106--2113. IEEE, 2009.

\bibitem{Karpathy_CVPR2015}
Andrej Karpathy and Li Fei-Fei.
\newblock Deep visual-semantic alignments for generating image descriptions.
\newblock In {\em Proceedings of the IEEE conference on computer vision and
  pattern recognition}, pages 3128--3137, 2015.

\bibitem{kendall2015bayesian}
Alex Kendall, Vijay Badrinarayanan, and Roberto Cipolla.
\newblock Bayesian segnet: Model uncertainty in deep convolutional
  encoder-decoder architectures for scene understanding.
\newblock {\em arXiv preprint arXiv:1511.02680}, 2015.

\bibitem{Kendall_NIPS2017}
Alex Kendall and Yarin Gal.
\newblock What uncertainties do we need in bayesian deep learning for computer
  vision?
\newblock In {\em Advances in neural information processing systems}, pages
  5574--5584, 2017.

\bibitem{Kendall_CVPR2018}
Alex Kendall, Yarin Gal, and Roberto Cipolla.
\newblock Multi-task learning using uncertainty to weigh losses for scene
  geometry and semantics.
\newblock 2018.

\bibitem{kim_NIPS2018bilinear}
Jin-Hwa Kim, Jaehyun Jun, and Byoung-Tak Zhang.
\newblock Bilinear attention networks.
\newblock In {\em Advances in Neural Information Processing Systems}, pages
  1571--1581, 2018.

\bibitem{Kim_ICLR2017}
Jin-Hwa Kim, Kyoung~Woon On, Woosang Lim, Jeonghee Kim, Jung-Woo Ha, and
  Byoung-Tak Zhang.
\newblock {Hadamard Product for Low-rank Bilinear Pooling}.
\newblock In {\em The 5th International Conference on Learning
  Representations}, 2017.

\bibitem{Kulkarni_CVPR2011}
Girish Kulkarni, Visruth Premraj, Sagnik Dhar, Siming Li, Yejin Choi,
  Alexander~C Berg, and Tamara~L Berg.
\newblock Baby talk: Understanding and generating image descriptions.
\newblock In {\em Proceedings of the 24th CVPR}. Citeseer, 2011.

\bibitem{kurmi2019curriculum}
Vinod~Kumar Kurmi, Vipul Bajaj, Venkatesh~K Subramanian, and Vinay~P
  Namboodiri.
\newblock Curriculum based dropout discriminator for domain adaptation.
\newblock {\em arXiv preprint arXiv:1907.10628}, 2019.

\bibitem{kurmi_cvpr2019attending}
Vinod~Kumar Kurmi, Shanu Kumar, and Vinay~P Namboodiri.
\newblock Attending to discriminative certainty for domain adaptation.
\newblock In {\em Proceedings of the IEEE Conference on Computer Vision and
  Pattern Recognition}, pages 491--500, 2019.

\bibitem{kurmi2019looking}
Vinod~Kumar Kurmi and Vinay~P Namboodiri.
\newblock Looking back at labels: A class based domain adaptation technique.
\newblock {\em arXiv preprint arXiv:1904.01341}, 2019.

\bibitem{Li_NIPS2016}
Ruiyu Li and Jiaya Jia.
\newblock Visual question answering with question representation update (qru).
\newblock In {\em Advances in Neural Information Processing Systems}, pages
  4655--4663, 2016.

\bibitem{Lu_NIPS2016}
Jiasen Lu, Jianwei Yang, Dhruv Batra, and Devi Parikh.
\newblock Hierarchical question-image co-attention for visual question
  answering.
\newblock In {\em Advances In Neural Information Processing Systems}, pages
  289--297, 2016.

\bibitem{malinin2018predictive}
Andrey Malinin and Mark Gales.
\newblock Predictive uncertainty estimation via prior networks.
\newblock In {\em Advances in Neural Information Processing Systems}, pages
  7047--7058, 2018.

\bibitem{Malinowski_NIPS2014}
Mateusz Malinowski and Mario Fritz.
\newblock A multi-world approach to question answering about real-world scenes
  based on uncertain input.
\newblock In {\em Advances in Neural Information Processing Systems (NIPS)},
  2014.

\bibitem{Noh_CVPR2016}
Hyeonwoo Noh, Paul Hongsuck~Seo, and Bohyung Han.
\newblock Image question answering using convolutional neural network with
  dynamic parameter prediction.
\newblock In {\em Proceedings of the IEEE Conference on Computer Vision and
  Pattern Recognition}, pages 30--38, 2016.

\bibitem{Patro_CVPR2018dvqa}
Badri Patro and Vinay~P. Namboodiri.
\newblock Differential attention for visual question answering.
\newblock In {\em The IEEE Conference on Computer Vision and Pattern
  Recognition (CVPR)}, June 2018.

\bibitem{Patro_EMNLP2018MDN}
Badri~Narayana Patro, Sandeep Kumar, Vinod~Kumar Kurmi, and Vinay Namboodiri.
\newblock Multimodal differential network for visual question generation.
\newblock In {\em Proceedings of the 2018 Conference on Empirical Methods in
  Natural Language Processing}, pages 4002--4012. Association for Computational
  Linguistics, 2018.

\bibitem{Patro_COLING2018learning}
Badri~Narayana Patro, Vinod~Kumar Kurmi, Sandeep Kumar, and Vinay Namboodiri.
\newblock Learning semantic sentence embeddings using sequential pair-wise
  discriminator.
\newblock In {\em Proceedings of the 27th International Conference on
  Computational Linguistics}, pages 2715--2729, 2018.

\bibitem{Ren_NIPS2015}
Mengye Ren, Ryan Kiros, and Richard Zemel.
\newblock Exploring models and data for image question answering.
\newblock In {\em Advances in Neural Information Processing Systems (NIPS)},
  pages 2953--2961, 2015.

\bibitem{selvaraju2017grad}
Ramprasaath~R Selvaraju, Michael Cogswell, Abhishek Das, Ramakrishna Vedantam,
  Devi Parikh, and Dhruv Batra.
\newblock Grad-cam: Visual explanations from deep networks via gradient-based
  localization.
\newblock In {\em Proceedings of the IEEE International Conference on Computer
  Vision (ICCV)}, 2017.

\bibitem{Shih_CVPR2016}
Kevin~J Shih, Saurabh Singh, and Derek Hoiem.
\newblock Where to look: Focus regions for visual question answering.
\newblock In {\em Proceedings of the IEEE Conference on Computer Vision and
  Pattern Recognition}, pages 4613--4621, 2016.

\bibitem{Singh_2019_CVPR}
Pravendra Singh, Vinay~Kumar Verma, Piyush Rai, and Vinay~P. Namboodiri.
\newblock Hetconv: Heterogeneous kernel-based convolutions for deep cnns.
\newblock In {\em The IEEE Conference on Computer Vision and Pattern
  Recognition (CVPR)}, June 2019.

\bibitem{smith2018understanding}
Lewis Smith and Yarin Gal.
\newblock Understanding measures of uncertainty for adversarial example
  detection.
\newblock {\em arXiv preprint arXiv:1803.08533}, 2018.

\bibitem{Socher_TACL2014}
Richard Socher, Andrej Karpathy, Quoc~V Le, Christopher~D Manning, and Andrew~Y
  Ng.
\newblock Grounded compositional semantics for finding and describing images
  with sentences.
\newblock {\em Transactions of the Association of Computational Linguistics},
  2(1):207--218, 2014.

\bibitem{srivastava2014dropout}
Nitish Srivastava, Geoffrey Hinton, Alex Krizhevsky, Ilya Sutskever, and Ruslan
  Salakhutdinov.
\newblock Dropout: a simple way to prevent neural networks from overfitting.
\newblock {\em The Journal of Machine Learning Research}, 15(1):1929--1958,
  2014.

\bibitem{teye2018bayesian}
Mattias Teye, Hossein Azizpour, and Kevin Smith.
\newblock Bayesian uncertainty estimation for batch normalized deep networks.
\newblock {\em arXiv preprint arXiv:1802.06455}, 2018.

\bibitem{Vinyals_CVPR2015}
Oriol Vinyals, Alexander Toshev, Samy Bengio, and Dumitru Erhan.
\newblock Show and tell: A neural image caption generator.
\newblock In {\em Proceedings of the IEEE Conference on Computer Vision and
  Pattern Recognition}, pages 3156--3164, 2015.

\bibitem{Xiong_arXiv2016}
Caiming Xiong, Stephen Merity, and Richard Socher.
\newblock Dynamic memory networks for visual and textual question answering.
\newblock In {\em Proceedings of International Conference on Machine Learning
  (ICML)}, 2016.

\bibitem{Xu_ECCV2016}
Huijuan Xu and Kate Saenko.
\newblock Ask, attend and answer: Exploring question-guided spatial attention
  for visual question answering.
\newblock In {\em European Conference on Computer Vision}, pages 451--466.
  Springer, 2016.

\bibitem{Xu_ICML2015}
Kelvin Xu, Jimmy Ba, Ryan Kiros, Kyunghyun Cho, Aaron Courville, Ruslan
  Salakhudinov, Rich Zemel, and Yoshua Bengio.
\newblock Show, attend and tell: Neural image caption generation with visual
  attention.
\newblock In {\em International Conference on Machine Learning}, pages
  2048--2057, 2015.

\bibitem{Yan_ECCV2016}
Xinchen Yan, Jimei Yang, Kihyuk Sohn, and Honglak Lee.
\newblock Attribute2image: Conditional image generation from visual attributes.
\newblock In {\em European Conference on Computer Vision}, pages 776--791.
  Springer, 2016.

\bibitem{Yang_CVPR2016}
Zichao Yang, Xiaodong He, Jianfeng Gao, Li Deng, and Alex Smola.
\newblock Stacked attention networks for image question answering.
\newblock In {\em Proceedings of the IEEE Conference on Computer Vision and
  Pattern Recognition}, pages 21--29, 2016.

\bibitem{zhang_ICLR2018learning}
Yan Zhang, Jonathon Hare, and Adam Pr{\"u}gel-Bennett.
\newblock Learning to count objects in natural images for visual question
  answering.
\newblock 2018.

\bibitem{Zhu_CVPR2016}
Yuke Zhu, Oliver Groth, Michael Bernstein, and Li Fei-Fei.
\newblock Visual7w: Grounded question answering in images.
\newblock In {\em Proceedings of the IEEE Conference on Computer Vision and
  Pattern Recognition}, pages 4995--5004, 2016.

\end{thebibliography}
}

%%%%%%%%%%%%%%%%%%%%%%%%%%%%%%%%%%%%%%%%%%%%%%%%%%%%%%%%%%
\section{supplementary}
\appendix
%%%%%%%%%%%%%%%%%%%%%%%%%%%%%%%%%%%%%%%%%%%%%%%%%%%%%%%%%%
This is the supplementary material for the paper `\textit{U-CAM: Visual Explanation using Uncertainty based Class Activation Maps}'. %We provide relevant additional qualitative and quantitative insights along with detailed ablation analysis for various aspects of the paper. 

\section{Questions for Figure 5(c, d)}

Table~\ref{tab_Uncertainty_measure_wh} and ~\ref{tab_Uncertainty_measure_yesno} have the questions  corresponding to ID's present in the Figure 5 of the main paper.
\begin{table}[!htb]
	\centering
	\begin{tabular}{|p{4.2cm}|c|}
		\hline
		\textbf{Question}  & \textbf{ID} \\
		\hline
		What does the person in this picture have on his face?     & 1\\
		
		How many baby elephants are there?    & 2\\
		What is in the bowl?    & 3\\
		Is the television on or off?    & 4\\     
		What color is the walk light?    & 5\\    
		Which way is its head turned?    & 6\\    
		How many people are riding on each bike?   & 7\\    
		What animal is in this picture?    & 8\\  
		What color is the road?   & 9\\        
		What color is the boy's hair?    & 10\\        \hline
		%   CL: category loss  , DL: depression  loss, VL: variation Loss
	\end{tabular}
	\vspace{0.1cm}
	\caption{\label{tab_Uncertainty_measure_wh} Reference for the Figure 5(c) of the main paper.}
\end{table}

\begin{algorithm}[h]
	\caption{Rank Correlation Procedure}\label{Correlation}
	\begin{algorithmic}[1]
		\Procedure{:}{Initialization}
		\State $P_{H}$: Probability distribution of Human Attention Map
		\State $P_{M}$	: Probability distribution of our model
		\BState \emph{\textbf{Rank}}:
		\State Compute Rank of $P_{H}$ : $R_{H}$
		\State Compute Rank of $P_{M}$ : $R_{M}$ 
		
		\BState \emph{\textbf{Rank Difference }}:
		
		\State Compute difference in rank between $R_{H}$ \& $R_{M}$ : $Rank_{Diff}$
		\State Compute square of rank difference $Rank_{Diff}$ :$S_{Rank\_Diff}$

		\BState \emph{\textbf{Rank Correlation}}:
		\State Compute Dimension of $P_{M}$ :N
		\State Compute Rank Correlation using :
		\[R_{Cor}= 1- {\frac{6*S_{Rank\_Diff}}{N^3 - N}}\]
		\EndProcedure
	\end{algorithmic}
\end{algorithm}

\begin{table}
	\centering
	\begin{tabular}{|p{4.2cm}|c|}
		\hline
		\textbf{Question}  & \textbf{ID} \\
		\hline
		Is this wheat bread?     & 1\\
		Is the cat looking at the camera?  & 2\\
		Is this chair broken?    & 3\\
		Are these animals monitored?    & 4\\     
		Does the cat recognize someone?    & 5\\    
		Is the figurine life size?   & 6\\    
		Is the smaller dog on a leash?   & 7\\    
		Is this in the mountains?   & 8\\  
		Is the woman sitting on the bench?   & 9\\        
		Is the church empty?    & 10\\        \hline
	\end{tabular}
	\vspace{0.1cm}
	\caption{\label{tab_Uncertainty_measure_yesno} Reference for the Figure 5(d) in the main paper.}
\end{table}

\section{Evaluation Methods}
\textbf{Accuracy:}VQA dataset has 3 type of answers: \textit{yes/no}, \textit{number} and \textit{other}. The evaluation is carried out using two test splits, i.e test-dev and test-standard. The question in corresponding test split are of two types: Open-Ended and Multiple-choice. Our model generates a single word answer on the open ended task. For each question there are 10 candidate answer provided with their respective confidence level. This answer can then be evaluated using accuracy metric defined as follows:

\begin{equation} 
Acc=\frac{1}{N}\sum_{i=1}^{N}{min\big(\frac{\sum_{t \in T^{i} }{\mathbb {I}[a_i=t]}}{3},1 \big)}
\end{equation}
Where $a_i$ the predicted answer and t is the annotated answer in the target answer set $T^{i}$  of the $i^{th}$ example and $\mathbb {I}[.]$ is the indicator function.
The predicted answer $a_i$ is correct if at least 3 annotators agree on the predicted answer. If the predicted answer is not correct then the accuracy score depends on the number of annotator that agree on the answer. Before checking accuracy, we need to convert the predicted answer to lowercase, number to digits and punctuation \& article to be removed. 
\textbf{Rank Correlation}: The detailed algorithm is given in Algorithm~\ref{Correlation}.
\\
	
\end{document}